\documentclass[journal]{IEEEtran}  
\pdfoutput=1
\usepackage{xparse} 
\usepackage{amsmath} 
\usepackage{amssymb}
\usepackage{amsfonts} 
\usepackage{dsfont}
\usepackage{bbm}
\usepackage{color}
\usepackage{verbatim}
\usepackage{multirow}
\usepackage[flushleft]{threeparttable}
\usepackage[font=small]{caption}
\usepackage{balance}
\usepackage{cleveref}
\usepackage{float}
\usepackage{stfloats}
\usepackage{cuted}
\usepackage{mathrsfs}
\voffset=\baselineskip
\usepackage{url}
\usepackage{framed}
\usepackage{booktabs}
\usepackage{arydshln}
\newcommand\T{\rule{0pt}{2.6ex}}        
\newcommand\B{\rule[-1.2ex]{0pt}{0pt}}  
\newcommand{\norm}[1]{\left\Vert#1\right\Vert}

\newcommand{\mbf}[1]{\mathbf{#1}}
\newcommand{\bbm}{\begin{bmatrix}}
\newcommand{\ebm}{\end{bmatrix}}

\usepackage{cite}
\usepackage[pdftex]{graphicx}
\usepackage{mathtools, nccmath}
\usepackage{multirow}
\usepackage{caption}
\usepackage[font=small]{caption}
\usepackage[font=small,position=b]{subcaption}
\usepackage{makecell}
\usepackage{url}

\begin{document}
\title{Robust Data-Driven Zero-Velocity Detection for  Foot-Mounted Inertial Navigation}

\author{Brandon Wagstaff,~\IEEEmembership{Member,~IEEE,}
		Valentin Peretroukhin,~\IEEEmembership{Member,~IEEE,}
        and~Jonathan~Kelly,~\IEEEmembership{Senior~Member,~IEEE}
\thanks{All authors are with the Space \& Terrestrial Autonomous Robotic Systems (STARS) Laboratory at the University of Toronto Institute for Aerospace Studies (UTIAS), Toronto, Ontario, Canada, M3H~5T6. Email: \texttt{<first name>.<last name>@robotics.utias.utoronto.ca}}
}

\IEEEoverridecommandlockouts
\IEEEpubid{\makebox[\columnwidth]{\copyright2019 IEEE \hfill} \hspace{\columnsep}\makebox[\columnwidth]{ }}

\maketitle
\IEEEpubidadjcol
\begin{abstract}
We present two novel techniques for detecting zero-velocity events to improve foot-mounted inertial navigation. Our first technique augments a classical zero-velocity detector by incorporating a motion classifier that adaptively updates the detector's threshold parameter. Our second technique uses a long short-term memory (LSTM) recurrent neural network to classify zero-velocity events from raw inertial data, in contrast to the majority of zero-velocity detection methods that rely on basic statistical hypothesis testing. We demonstrate that both of our proposed detectors achieve higher accuracies than existing detectors for trajectories including walking, running, and stair-climbing motions. Additionally, we present a straightforward data augmentation method that is able to extend the LSTM-based model to different inertial sensors without the need to collect new training data.
\end{abstract}


\IEEEpeerreviewmaketitle

\vspace*{-2mm}
\section{Introduction}
\IEEEPARstart{R}{eliable} localization forms the basis of many mobile autonomy systems. Although still in its nascent stages, the transfer of these algorithms to the domain of human localization may prove vital for coordinating personnel in chaotic emergency environments. For teams of first responders such as firefighters, for example, the ability to track team members' locations in real time can facilitate more effective response strategies and expedite personnel extraction in case of unexpected events or injuries \cite{Fischer2010}.

While Global Navigation Satellite System (GNSS) receivers offer an attractive localization solution for outdoor environments, their limited accuracy indoors---largely due to signal absorption and spurious reflections---makes their sole use impractical for first response. For reliable indoor navigation, many researchers \cite{Foxlin:2005,Godha:2008,Nilsson:2014,Jimenez:2010} have instead proposed the use of body-mounted inertial measurement units (IMUs) to estimate an individual's motion relative to a known origin. These sensors are lightweight, low power, independent of any external infrastructure, and straightforward to deploy. A zero-velocity-aided inertial navigation system (INS) is a well-studied and popular solution for IMU-based pedestrian localization \cite{Foxlin:2005,Nilsson:2014}. During bipedal locomotion, a foot-mounted IMU repeatedly passes through a stationary phase (relative to a fixed reference frame), at which time the pose estimation error can be corrected by applying a `zero-velocity update.' If correctly identified, zero-velocity updates can significantly improve localization estimates. However, false-positive detections cause the length of the wearer's trajectory to be underestimated, while false-negative detections (i.e., missed detections) lead to rapid and unbounded error growth.

Robust zero-velocity detection is an open research problem; significant advances must be made before existing systems are capable of high-accuracy localization during general locomotion involving periods of walking, running, and stair-climbing (among other possible motions). In past work, we developed two robust zero-velocity detectors that operate consistently across multiple motion types. In \cite{Wagstaff:2017}, we introduced a learning-based motion classifier that identifies a wearer's motion type and uses this information to adjust the threshold of a `classical' zero-velocity detector to be optimal for the current motion. In \cite{Wagstaff:2018}, we trained a long short-term memory (LSTM) neural network \cite{Hochreiter:1997} to directly classify zero-velocity events, effectively replacing the classical detector entirely and obviating the need for threshold tuning. 

Herein, we substantially expand on our prior results by providing a quantitative comparison of the proposed zero-velocity detectors with existing methods. Additionally, we present a new stair-climbing dataset that includes ground truth position data, and we quantitively demonstrate that our proposed detectors dramatically improve the vertical motion estimates of a zero-velocity-aided INS compared to existing detectors. To the best of the authors' knowledge, no systematic study of stair-climbing has appeared previously in the literature. In short, the novel contributions of this work are:
\begin{enumerate}
	\item the extension of our motion classifier to include a third stair-climbing class,
	\item a direct comparison of our two state-of-the-art zero-velocity detectors with existing detectors on two datasets, where extensive efforts have been made to obtain accurate ground truth,
	\item the introduction of a new technique for retraining the LSTM-based zero-velocity classifier to work with different IMUs without requiring the recollection of training data, and
	\item the release of a publicly available zero-velocity-aided INS called PyShoe\footnote{See \url{https://github.com/utiasSTARS/pyshoe}} and a mixed-motion dataset that includes ground truth position information.	
\end{enumerate}

\section{Background}
\label{sec:related-work}
In a standard strapdown INS, the IMU measures linear accelerations and angular velocities, which are then rotated into the navigation frame and integrated to yield pose updates as a function of time. With relatively inexpensive inertial sensors that employ  microelectromechanical systems (MEMS), this approach is accurate for only short durations, as sensor biases and noise quickly lead to estimates of progressively poorer quality. For dead reckoning alone, position error grows cubically with time \cite{ZVDetect} and some form of additional correction is necessary for accurate long-duration inertial odometry. To combat inertial sensor drift, one attractive technique (that does not require external aiding or extant maps) is to mount the IMU on the foot of an individual and rely on zero-velocity updates. These updates are \textit{pseudo-measurements} of the velocity state that occur during midstance, that is, the portion of the human gait during which the foot is flat on the ground and stationary relative to the navigation frame. By incorporating such pseudo-measurements into the INS through a Bayesian filter such as an extended Kalman filter (EKF), dead reckoning is limited to the intervals between footfalls, as opposed to over the entire tracked motion \cite{Foxlin:2005}.

\begin{figure}[t]
	\centering
	\includegraphics[width=0.9\columnwidth]{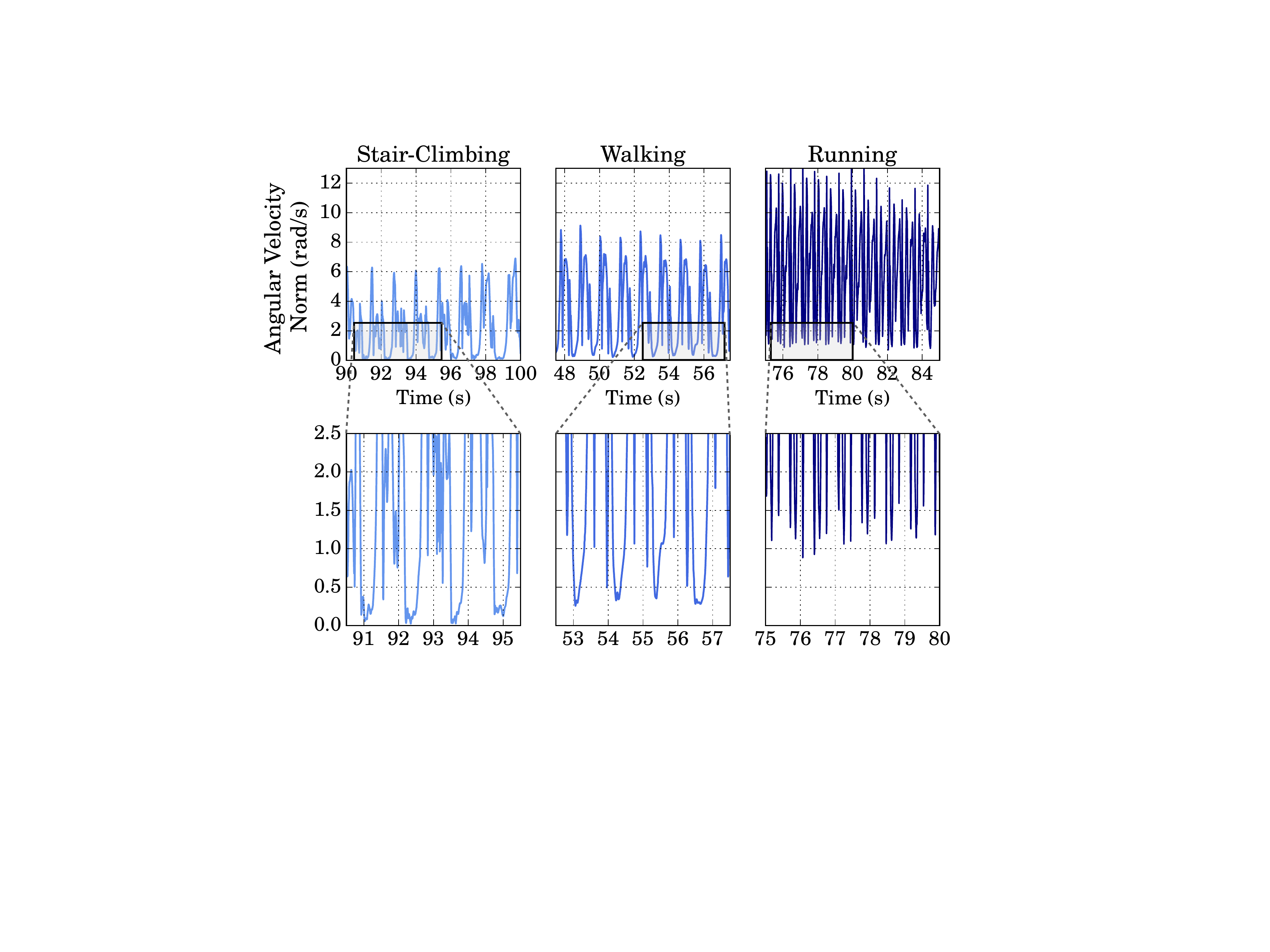}
	\caption{The norm of the angular velocity vector of a foot-mounted IMU during various motions. The angular velocity profile is highly dependent on the motion type.}
	\label{fig:ang-vel-plot}
	\vspace{-0.2cm}
\end{figure}

The typical method of zero-velocity detection is a likelihood ratio test (LRT) that uses the IMU readings to estimate the point(s) in time at which the foot is stationary \cite{ZVDetect}. The LRT is typically a heuristic indicator of two conditions: (1) that the measured IMU linear acceleration is equal in magnitude and opposite in direction to the local gravity vector, and (2) that the norm of the measured angular velocity is zero. In practice, such conditions will never be \textit{exactly} fulfilled; rather, these heuristics must be encoded in a quantifiable metric to which a thresholding procedure can be applied. The threshold parameter is typically tuned for a given dataset by selecting the value that leads to the minimum position error \cite{zupteval} and then is subsequently fixed during future operation. However, a fixed threshold is only suitable for uniform motions, which limits the detector's effectiveness during motions that are `dynamic' in nature.\footnote{Here, we define dynamic motions to be those which change in type (e.g., walking to running, to stair-climbing, to crawling, etc.), or intensity (e.g., slow walking versus fast walking, and jogging versus sprinting).} Dynamic motions result in varied acceleration and angular rate profiles when the foot makes contact with the ground (see \Cref{fig:ang-vel-plot} for an illustration of angular rate variations as a function of motion type and \cite{chan:1994} for an analysis of impact forces during locomotion). Since zero-velocity detection heuristics are based on these quantities, it becomes difficult to tune a single fixed-threshold detector to operate reliably throughout dynamic motions.

A range of statistical tests for zero-velocity detection have been proposed and have been studied by Skog et al.\ \cite{zupteval,ZVDetect} and Olivares et al.\ \cite{Olivares:2012}. The detectors that rely on these tests all use fixed thresholds that are susceptible to failures during dynamic motions. A threshold that is too low leads to a failure to report detections when the foot is in fact stationary (i.e., false negatives), while a threshold that is too high leads to false reports of zero-velocity events when the foot is moving (i.e., false positives). Both cases result in irreversible error accumulation \cite{Nilsson:2012}. To improve upon fixed-threshold zero-velocity detectors, several methods in the literature have attempted to set a varying threshold that adapts to the wearer's motion. For example, by assuming a Bayesian detection model, the threshold can be factored into a time-varying prior on the zero-velocity hypothesis and a time-varying loss for missed detections \cite{Wahlstrom:2019}. Alternatively, the threshold can be modelled explicitly as a function of gait frequency \cite{Sensors2016_2}, estimated linear velocity \cite{context-adaptive,Sensors2016}, or estimated angular velocity \cite{Ma:2017}.

However, developing a hand-crafted model that applies a proper threshold for \textit{all} types of motion remains challenging. In our work, we attempt to remedy this through the use of data-driven models that \textit{learn} the relationship between raw inertial data and latent variables such as the wearer's motion type or the velocity state of the IMU. In line with our approach, others \cite{Park2016,Rantanen:2018} have used data-driven motion classification to adaptively update the zero-velocity threshold. Our motivation for the use of learning stems in part	 from other work that has successfully applied learning-based methods to process inertial data. For example, Hannink et al.\ \cite{Hannink:2018} presented a method to train a deep convolutional neural network (CNN) to predict human stride length from inertial measurements, Chen et al.\ \cite{Chen:2018} used an LSTM network to directly estimate a trajectory from raw inertial data, and Cort\'es et al.\ \cite{Cortes:2018} trained a CNN to regress velocities from IMU outputs to improve trajectory estimates.

\section{Approach}

\begin{figure}[t]
	\centering
	\includegraphics[width=0.85\columnwidth]{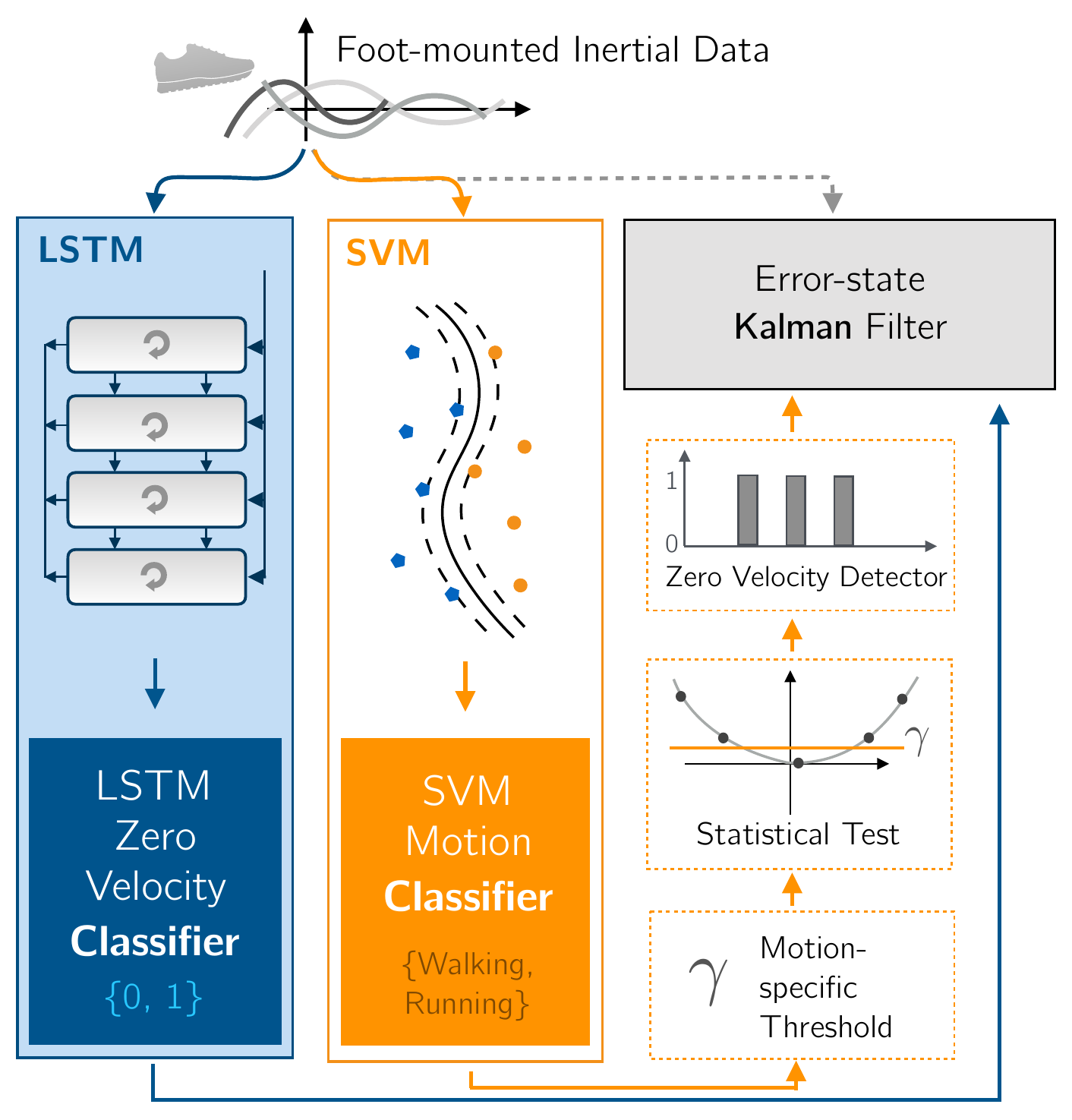}
	\caption{Our localization pipeline uses learning-based classifiers to improve zero-velocity detection: the motion classifier actively updates the threshold of an existing zero-velocity detector to be optimal for the current motion type, while the zero-velocity classifier replaces the zero-velocity detector with a learned long short-term memory (LSTM) network.}
	\label{fig:system_overview}
	\vspace{-0.3cm}
\end{figure}

To create robust zero-velocity detectors, our approach eschews threshold regression in favour of learned classifiers, which have achieved state-of-the-art results in many domains. We present two classification-based methods for zero-velocity detection and describe straightforward ways to generate training labels for each that facilitate supervised learning. First, we train an SVM to classify a wearer's motion type from a short sequence of IMU data; given the known motion type, we update the threshold of an existing zero-velocity detector to a pre-defined value optimized for that motion type. Second, we adopt a purely learning-based strategy by training a recurrent neural network to directly classify when the IMU is stationary. Given a sequence of inertial data, our network outputs a binary prediction of the velocity state of the IMU (i.e., whether the IMU is stationary or moving). The network is trained with a loss function that compares the network prediction to zero-velocity ground truth. Both approaches, illustrated in \Cref{fig:system_overview}, are data-driven, as they augment or replace the classical zero-velocity detector with models that are learned.

\subsection{Zero-Velocity Aided INS}
\label{sec:ins}

Our system uses an error-state (extended) Kalman filter (ESKF) to estimate the wearer's motion from inertial data. The filter state at time step $k$ consists of the IMU's position, $\mbf p_k$, velocity, $\mbf v_k$, and orientation (parameterized as a unit-quaternion), $\mbf q_k$, all expressed with respect to a fixed local navigation frame,
\begin{align}
\label{eq:state}
\mathbf{x}_k &= \bbm \mbf p^{T}_k & \mbf v^{T}_k &  \mbf q^{T}_k \ebm^T.
\end{align}  
The nominal state is propagated forward using a discrete-time nonlinear motion model \cite{Nilsson:2014},
\begin{align}
\label{eq:naveq}
\mbf{x}_k = \bbm
\mbf{p}_k\\
\mbf{v}_k\\
\mbf{q}_k\\ \ebm 
= 
\bbm
\mbf{p}_{k-1} + \mbf{v}_{k-1}\Delta t \\ 
\mbf{v}_{k-1} + \left( \mbf{R}(\mbf{q}_{k-1})\,\mbf{a}_k + \mbf{g} \right)\Delta t  \\
\mbf{\Omega} (\boldsymbol{\omega}_k \Delta t)\,\mbf{q}_{k-1} \\
\ebm,
\end{align}
where $\mbf q_{k-1}$, the orientation quaternion, is updated with the incremental IMU rotation over the interval $\Delta t$ by integrating the angular velocity $\boldsymbol{\omega}_k$ (making use of the quaternion update matrix $\mbf{\Omega}(\cdot)\in\mathds{R}^{4\times4}$). The velocity, $\mbf v_{k-1}$, is updated with the integrated linear acceleration, $\mbf a_k$, after expressing the acceleration vector in the navigation frame and removing the contribution due to gravity.\footnote{Here, $\mathbf{R}(\cdot)$ converts a unit quaternion to the corresponding rotation matrix.} Lastly, the position, $\mbf p_{k-1}$, is updated with the integrated velocity term. Note that we apply the first-order Euler integration method, but that other higher-order integration schemes could also be used. This model does not incorporate accelerometer or gyroscope biases (although they are observable through zero-velocity detection) because, as discussed in \cite{Nilsson:2012}, inaccuracies resulting from the zero-velocity assumption are the dominant error source (i.e., we assume that the IMU is \textit{exactly} stationary during midstance, when in reality it always has a small, but non-zero, velocity). In parallel with the nominal state, the error state is propagated forward in time to track noise and perturbations to the system.

When a zero-velocity event is detected, the current velocity state is compared to zero to produce an estimate of the velocity error. We make the usual assumption that the zero-velocity pseudo-measurement is corrupted by additive Gaussian noise and apply the standard ESKF correction step. The reader is referred to \cite{Foxlin:2005,Sola:2016} (and to the documentation for our open source INS, PyShoe) for a thorough discussion of the ESKF and its application to zero-velocity-aided inertial navigation.

\subsection{Classical Fixed-Threshold Zero-Velocity Detectors}
\label{sec:detectors}

Our approach to zero-velocity detection leverages the accuracy of `classical' detectors for fixed motions (i.e., of the same type and intensity). The output of the SVM motion classifier is used to update the threshold of a classical detector, while the outputs of one or more classical detectors are used to generate labels for training of the LSTM network. Below, we provide brief descriptions of three of the five detectors used to train our classifiers.\footnote{We omit descriptions of the acceleration-moving variance detector (AMVD) \cite{ZVDetect} and the memory-based graph theoretic detector (MBGTD) \cite{Olivares:2012} due to space constraints, and refer the reader to the respective papers.}

\subsubsection{Stance Hypothesis Optimal Estimation (SHOE) Detector}

The SHOE detector \cite{ZVDetect} uses a small temporal window of IMU readings to either accept or reject the hypothesis that the sensor is moving relative to the navigation frame. This hypothesis is rejected (and the alternate hypothesis, that the IMU is stationary, is accepted) when the average of the $\ell^2$-norm of the gravity-subtracted acceleration, combined with an angular velocity term, falls below a threshold, $\gamma$.  The binary SHOE detector output is given by
\begin{align}
\label{eq:shoe}
y_k \hspace{-0.06cm} = \hspace{-0.06cm}\begin{cases}
1,  & \hspace{-0.2cm} \text{if } \frac{1}{N} \sum\limits_{n=k}^ {k+N-1}\hspace{-0.06cm}\left( \frac{1}{\sigma_a^2}\norm{ \mathbf{a}_n - g\frac{\bar{\mathbf{a}}}{\norm{\bar{\mathbf{a}}}}}^2 + \frac{1}{\sigma_\omega^2}\norm{\boldsymbol{\omega}_n}^2 \right)\hspace{-0.06cm} < \gamma, \\
0, & \hspace{-0.2cm} \text{otherwise}.
\end{cases}
\end{align}

\noindent where $\mbf a_n$, $\boldsymbol{\omega}_n \in\mathds{R}^{3}$ are linear acceleration and angular velocity samples, respectively, from a window of N samples. The two terms in the SHOE detector LRT are weighted by the variances of the linear acceleration and angular velocity measurements, $\sigma_a^2$ and $\sigma_\omega^2$, respectively. The term $\mbf {\bar{a}}$ is the per-channel mean of the linear acceleration over all of the samples within the current window.

\subsubsection{Angular Rate Energy Detector (ARED)}

The ARED \cite{ZVDetect} shares the same angular velocity detection component as the SHOE detector, but omits the linear acceleration term. This detector produces false positives when the foot moves without rotating,
	\begin{align}
	\label{eq:ARED}
	y_k = \begin{cases}
	1,  & \hspace{-0.2cm} \text{if } \frac{1}{N} \sum\limits_{n=k}^{k+N-1} \norm{\boldsymbol{\omega}_n}^2  < \gamma_\omega, \\
	0, & \hspace{-0.2cm} \text{otherwise}.
	\end{cases}
	\end{align}
Note that the trigger threshold, $\gamma_\omega$, is different than that of the SHOE detector.

\subsubsection{VICON-Based Zero-Velocity Detector}
VICON-based zero-velocity detection is an approach we developed in \cite{Wagstaff:2017}. A VICON motion tracking system is able to determine the position of the IMU accurately, such that the IMU velocity can be directly computed through numerical differentiation. The velocity norm (i.e., the foot speed) can be determined and a threshold $\gamma_v$ applied to detect when the foot velocity is nearly zero,
	\begin{align}
	\label{eq:vicon-zv}
	y_k = \begin{cases}
	1,  & \hspace{-0.2cm} \text{if } \norm{\frac{\mbf{p}_k - \mbf{p}_{k-1}}{t_k - t_{k-1}}} < \gamma_v, \\
	0, & \hspace{-0.2cm} \text{otherwise}.
	\end{cases}
	\end{align}

Our VICON-based detector plays an important role in identifying midstance when the wearer's foot is moving directly upwards or downwards (e.g., when stair-climbing). During such a motion, the angular velocity and linear acceleration of the foot are both small, which can cause classical detectors to report false positives. Since the VICON-based detector measures foot speed only, it is less likely to erroneously detect midstance during upwards or downwards motions. With a nearly perfect velocity estimate, one would expect the VICON-based detector to be more accurate than other detectors that rely on the IMU measurements only. However, the significant impact forces associated with locomotion can cause the VICON markers mounted to the IMU to vibrate, which results in a noisy VICON-based velocity estimate during walking and running.

\subsection{SVM Motion Classification for Adaptive Thresholding}
\label{sec:svm-classifier}

Our approach for motion-adaptive zero-velocity detection incorporates a motion classifier into the zero-velocity detection pipeline, as illustrated in \Cref{fig:system_overview}. Since the optimal zero-velocity threshold is highly dependent on motion type, we determine a motion-specific threshold for each motion type and use the motion classifier to adaptively update the threshold of a classical zero-velocity detector. In prior work \cite{Wagstaff:2017}, we trained a two-class SVM motion classifier to distinguish walking from running. Herein, we build on our prior work by incorporating a stair-climbing class, which improves the system's localization accuracy for movement within multi-storey buildings. To train the SVM we use a dataset that is labelled according to the test subject's known motion type. We select a specific threshold for each motion, and use the motion classifier to actively change the zero-velocity threshold of the SHOE detector (see Equation \ref{eq:shoe}) to be optimal for the current motion type; a low threshold is assigned for walking and stair-climbing and a high threshold is assigned for running (according to \Cref{fig:ang-vel-plot}). Throughout this paper, the term ``motion-adaptive zero-velocity detector'' is used to refer to the SHOE detector in combination with our SVM motion classifier.

\subsection{LSTM-Based Zero-Velocity Classification}
\label{sec:lstm-zv}

Our second learning-based approach to zero-velocity detection removes most modelling assumptions (e.g., the assumption of distinct motion classes) by entirely replacing the zero-velocity detector with a recurrent neural network (RNN). We use a supervised learning approach to train a binary classifier for zero-velocity detection from a dataset containing inertial measurements with zero-velocity labels. Inertial measurements are sequential and low-dimensional (six values per time step), which makes them well suited for RNN sequence learning. In contrast to classical zero-velocity detection methods that utilize a short sequence of data, an RNN is able to propagate its memory state across long input sequences. The network can therefore make use of temporal context during classification, and is able to exploit the periodic nature of the human gait. Our approach uses a type of RNN called a long short-term memory (LSTM) network, which is a popular choice for sequential data processing; an LSTM network is able to efficiently backpropagate gradient information while being resistant to vanishing or exploding gradients (a known problem in many RNN models). The reader is referred to \cite{Lipton:2015} for a more detailed review of the theory of LSTM networks.

We generate binary zero-velocity labels (moving versus stationary) for each inertial dataset and use the labels for supervised training of our zero-velocity classifier. While existing work has performed dataset labelling through pressure sensing \cite{zupteval} or manual annotation \cite{Olivares:2012}, we leverage the output of existing classical detectors that have been optimized to minimize the position error over a short movement sequence of one motion type. This zero-velocity labelling technique is based on the assumption that existing hand-crafted zero-velocity detectors can produce optimal outputs if two conditions are met: the wearer's motion is of a fixed type and intensity, and the zero-velocity threshold is optimized for the current motion. For labelling, we make use of the five zero-velocity detectors mentioned in Section \ref{sec:detectors}; for every motion trial, the threshold of each detector is varied to identify the optimal detector-specific value. The output from the detector that produces the lowest position error over a trial is used as the approximate zero-velocity ground truth for the trial.

\section{Inertial Localization Experiments}
\label{sec:experiments}

In this section, we describe the data collection and training procedures for the proposed data-driven zero-velocity detectors. The training data were collected by a VICON infrared motion tracking system and the proposed methods were evaluated using two datasets: a hallway dataset consisting of walking and running motions and a novel stair-climbing dataset. \Cref{tab:datasets} summarizes these datasets.

\begin{table}[]
	\renewcommand{\arraystretch}{1.1}
	\centering
	\caption{An overview of the datasets discussed in this paper. The VICON dataset was used to train our learning-based classifiers. The remaining datasets were used for testing.} 
	\begin{threeparttable}
\begin{tabular}{c@{\hspace{0.9\tabcolsep}}c@{\hspace{0.9\tabcolsep}}c@{\hspace{0.9\tabcolsep}}c@{\hspace{0.9\tabcolsep}}@{\hspace{0.9\tabcolsep}}}
	\toprule
	\multirow{2}{*}{\begin{tabular}[c]{@{}c@{}}\textbf{Dataset}\\ (Relevant Section)\end{tabular}} & \multirow{2}{*}{\begin{tabular}[c]{@{}c@{}}\textbf{IMU}\\ (Frequency)\end{tabular}} & \multicolumn{2}{c}{\textbf{Ground Truth}} \\ \cmidrule{3-4} 
	&  & Method & Type \\ \midrule
	\begin{tabular}[c]{@{}c@{}}\textbf{VICON}$^\dagger$\\ (\Cref{sec:datacollection,sec:svmtrain,sec:lstmtrain}) \end{tabular}& \begin{tabular}[c]{@{}c@{}}\texttt{MicroStrain}\\ \texttt{3DM-GX3-25}\\ (200 Hz)\end{tabular} & \begin{tabular}[c]{@{}c@{}}Motion \\ Tracking\\ (200 Hz)\end{tabular} & 3D Position \\ \midrule
	\begin{tabular}[c]{@{}c@{}}\textbf{Hallway 1}$^\dagger$\\ (\Cref{sec:hallway-results})\end{tabular} & \begin{tabular}[c]{@{}c@{}}\texttt{VectorNav}\\\texttt{VN-100}\\ (200 Hz)\end{tabular} & \begin{tabular}[c]{@{}c@{}}Surveyed\\ (8 points)\end{tabular} & 3D Position \\ \midrule
	\begin{tabular}[c]{@{}c@{}}\textbf{Hallway 2}\\ (\Cref{sec:domain-adapt})\end{tabular} & \begin{tabular}[c]{@{}c@{}}\texttt{Osmium} \\ \texttt{MIMU22BTP}\\ (125 Hz)\end{tabular} & \begin{tabular}[c]{@{}c@{}}Surveyed\\ (6 points)\end{tabular} & 3D Position \\ \midrule
	\begin{tabular}[c]{@{}c@{}}\textbf{Stair-Climbing}$^\dagger$\\ (\Cref{sec:stair-results})\end{tabular} & \begin{tabular}[c]{@{}c@{}}\texttt{VectorNav}\\\texttt{VN-100} \\ (200 Hz)\end{tabular} & \begin{tabular}[c]{@{}c@{}}Surveyed \\ Floor Height\end{tabular} & \begin{tabular}[c]{@{}c@{}}1D \\ (Vertical Only)\end{tabular} \\ \bottomrule
\end{tabular}
\begin{tablenotes}
	\small
	\item $\dagger$ Publicly available in our open source repository.
\end{tablenotes}
\end{threeparttable}
\label{tab:datasets}
\vspace{-0.3cm}
\end{table}

\subsection{VICON Dataset Collection}
\label{sec:datacollection}

The VICON dataset consists of 60 motion trials (a distance of approximately 1 km in total). All trials were performed by a single person who had an IMU mounted to the top of their right foot, secured by  shoe laces. For the initial data collection, we used a LORD MicroStrain 3DM-GX3-25 IMU operating at 200 Hz. The motion trials primarily consist of walking and running at a range of speeds, but also include five stair-climbing trials and five crawling trials. Approximately ten trials involve irregular motions such as shuffling, backwards-walking, and vertical foot raising.  Each motion trial was performed in our 5 m $\times$ 5 m VICON motion capture area, where ground truth position updates at 200 Hz were available.  

We ensured that the motion within each trial consisted of a fixed type and gait frequency, which allowed for proper labelling of the data. To train the SVM motion classifier, we labelled all training data with the known motion type of the sequence. To train the LSTM zero-velocity classifier, binary zero-velocity labels were required, which we produced using the procedure described in \Cref{sec:lstm-zv}. In summary, we optimized each detector's threshold by minimizing the position error relative to ground truth, and selected the zero-velocity output from the best-performing detector. A line search was used to identify the threshold that minimized the average position error.

\begin{table}[b]
	\renewcommand{\arraystretch}{1.05}
    \vspace{-2mm}
	\centering
	\begin{threeparttable}
		\caption{Results for zero-velocity labelling of data from the 60 motion trials carried out within our VICON motion capture area. Num.\ Trials refers to the number of trials in which the particular detector produced the lowest ARMSE after threshold optimization.}
		\label{tab:vicon-trials}
		\begin{tabular*}{\columnwidth}{l@{\extracolsep{\fill}}*{5}{c}}
			\begin{tabular}[c]{@{}c@{}}Detector\end{tabular} & \begin{tabular}[c]{@{}c@{}}Avg.\\ Error (m)\end{tabular}  & \begin{tabular}[c]{@{}c@{}}Min.\\ Thresh.\end{tabular} & \begin{tabular}[c]{@{}c@{}}Max.\\ Thresh.\end{tabular} & \begin{tabular}[c]{@{}c@{}}Num.\\ Trials\end{tabular} \\ \midrule \T \T \B
			VICON & 0.074          & $2.25\times10^{-2}$ & $8.25\times10^{-1}$ & 15          \\
			SHOE  & \textbf{0.068} & $4.75\times10^5$    & $6.50\times10^8$    & \textbf{30} \\
			AMVD  & 0.336           & $1.00\times10^{-3}$ & $1.95\times10^{0}$  & 0           \\
			ARED  & 0.075          & $1.25\times10^{-2}$ & $2.70\times10^{0}$  & 13          \\
			MBGTD & 0.329          & $5.75\times10^{-3}$ & $9.75\times10^{-1}$ & 2           \\ \bottomrule
		\end{tabular*}
		\begin{tablenotes}
			\small
			\item This table previously appeared in \cite{Wagstaff:2018}.
		\end{tablenotes}
	\end{threeparttable}
\end{table}

\Cref{tab:vicon-trials} details the number of times each detector (with its per-trial optimized threshold) resulted in the lowest error for one of the 60 trials. The SHOE detector produced the lowest average root-mean-square error (ARMSE) in general. The VICON-based detector and the ARED were comparable to the SHOE detector in terms of accuracy, while the MBGTD and the AMVD performed substantially worse. \Cref{tab:vicon-trials} also shows the threshold range that resulted in the minimum error over the motion trials. The wide range of the per-trial optimal threshold indicates how dependent the threshold parameter is on the motion type and is further evidence that a single threshold is not suitable for dynamic motions.

\subsection{SVM Training}
\label{sec:svmtrain}

We trained our motion-adaptive zero-velocity classifier using a subset of the VICON data (four walking trials and five running trials), along with six additional stair-climbing trials (see \Cref{sec:stair-results}). The first half of the data were used for training and the second half for validation. A total of 2000 samples were generated per trial, where each sample spanned 200 IMU time steps (all six IMU channels were concatenated, resulting in a sample size of 1200). Each sample was normalized (by scaling the acceleration and angular velocity channels to be of unit norm) and then randomly rotated to simulate an arbitrary IMU orientation. Every sample generated from a trial was labelled with the trial's known motion type.

\begin{figure}[]
	\small
	\centering
	\includegraphics[width=0.8\columnwidth]{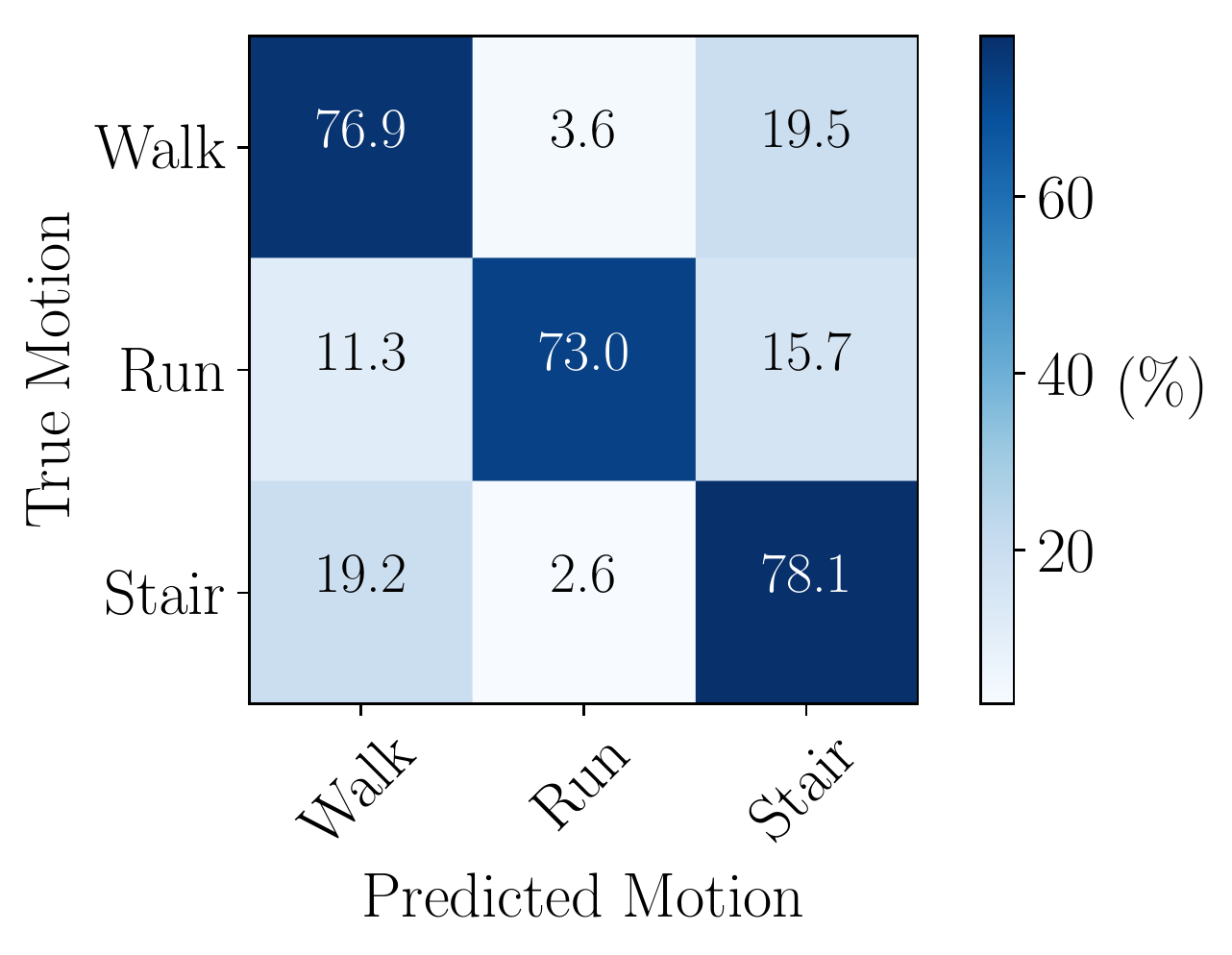}
	\caption{Confusion matrix illustrating the accuracy of our three-motion SVM classifier.}
	\label{fig:conf-matrix}
	\vspace{-0.3cm}
\end{figure}

The SVM was trained using the Python \texttt{scikit-learn} library \cite{scikit-learn} with the `one-against-one' approach for multi-class classification. We chose a radial basis function (RBF) kernel with a kernel coefficient of 0.001. \Cref{fig:conf-matrix} presents a confusion matrix indicating the accuracy of the motion classifier on the validation set. Our classifier achieved accuracies of over 70\% for all three motions, but often mis-classified stair-climbing as walking and vice versa. We attribute this to the similarity of walking and stair-climbing, and note that this is not detrimental to the accuracy of the system since the optimal thresholds for walking and stair-climbing are similar. 

\begin{figure*}
	\centering
	\begin{subfigure}[]{0.31\textwidth}
		\includegraphics[width=\textwidth]{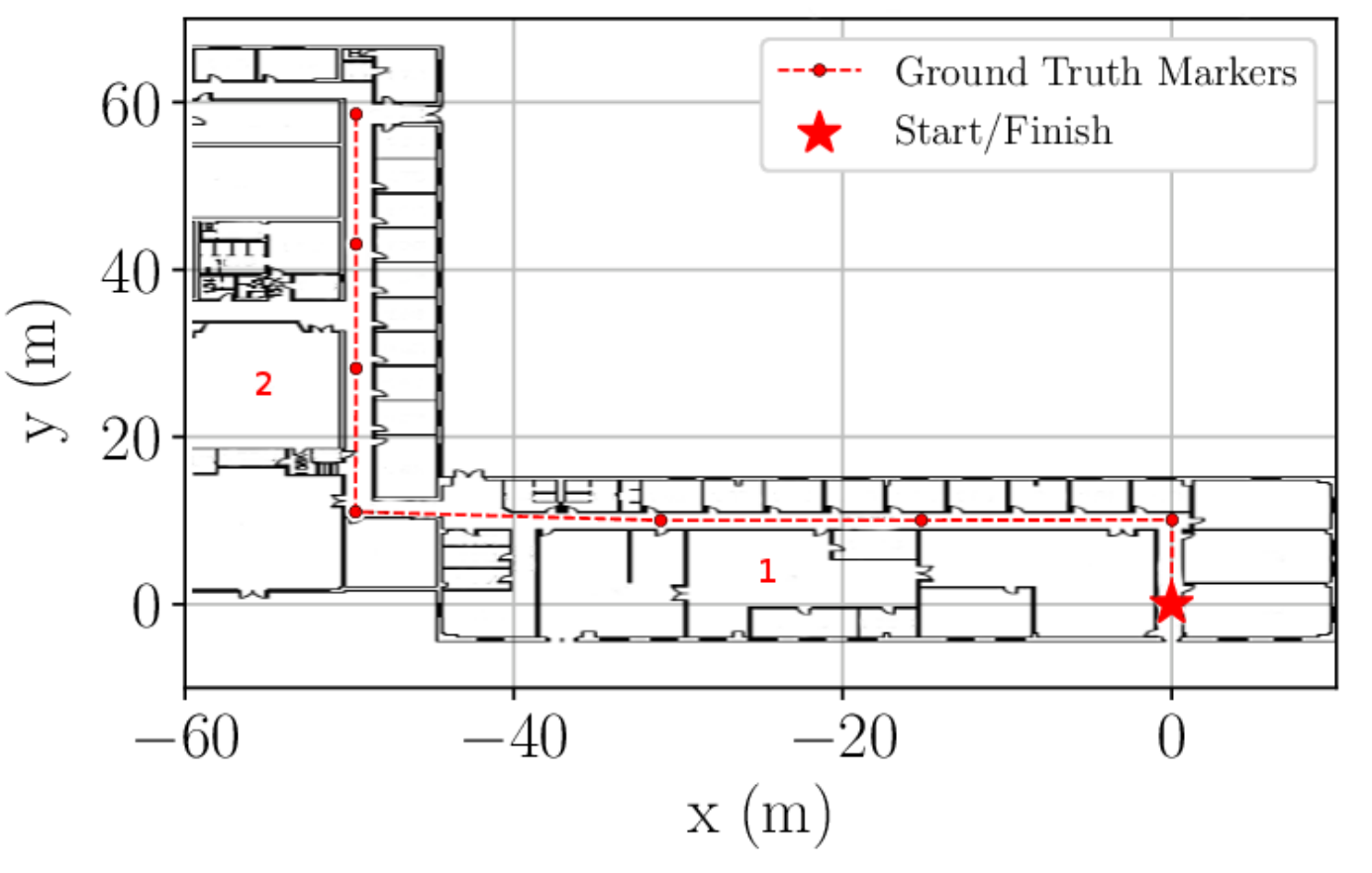}
		\caption{UTIAS test course.}
		\label{fig:hallway}
	\end{subfigure}
	\hfill
	\begin{subfigure}[]{0.31\textwidth}
		\includegraphics[width=\textwidth]{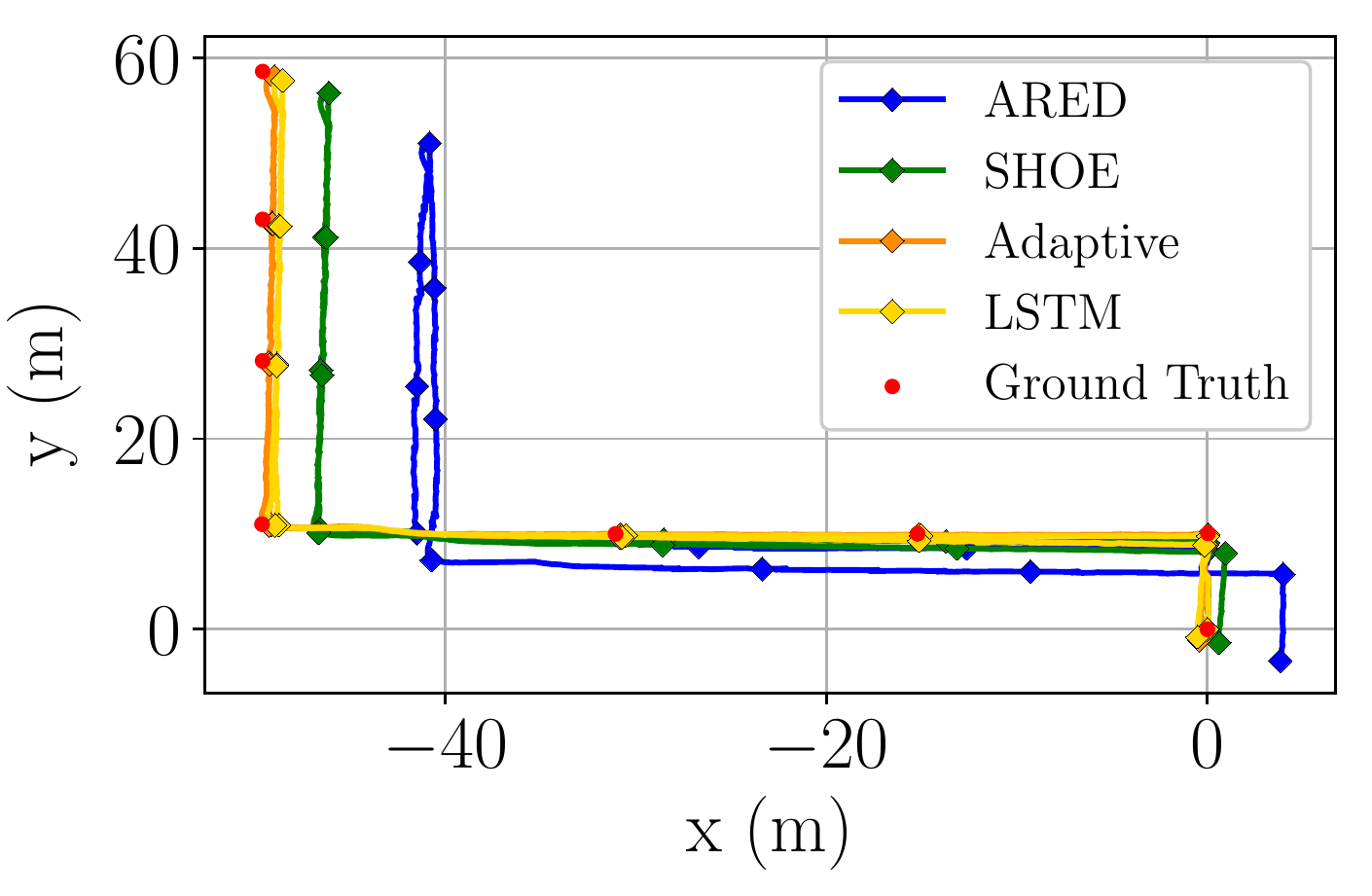}
		\caption{Walking trial.}
		\label{fig:hallwalk}
	\end{subfigure}
	\hfill
	\begin{subfigure}[]{0.31\textwidth}
		\includegraphics[width=\textwidth]{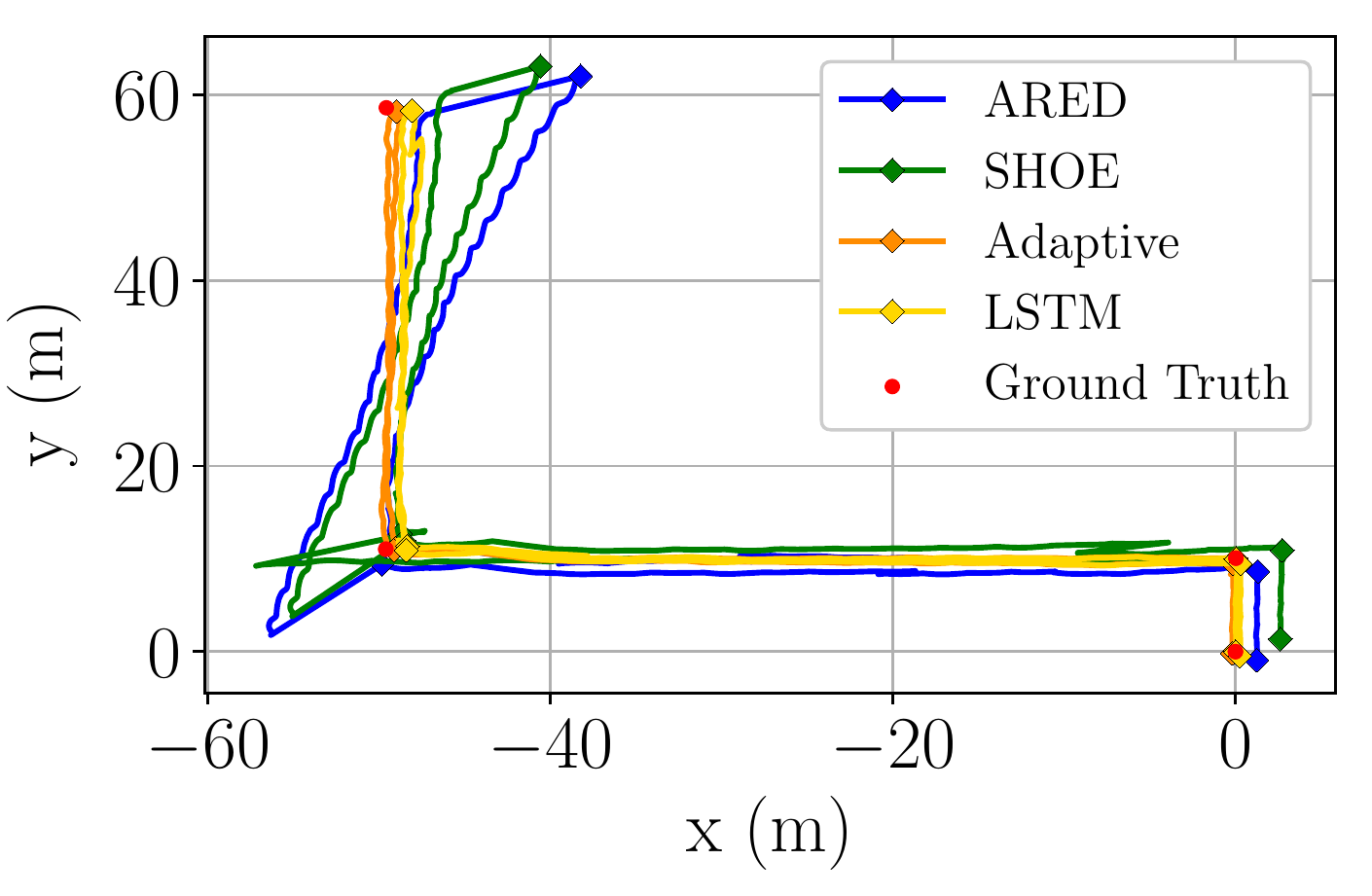}
		\caption{Running trial.}
		\label{fig:hallrun}
	\end{subfigure}\\
\hfill
	\begin{subfigure}[]{0.31\textwidth}
		\includegraphics[width=\textwidth]{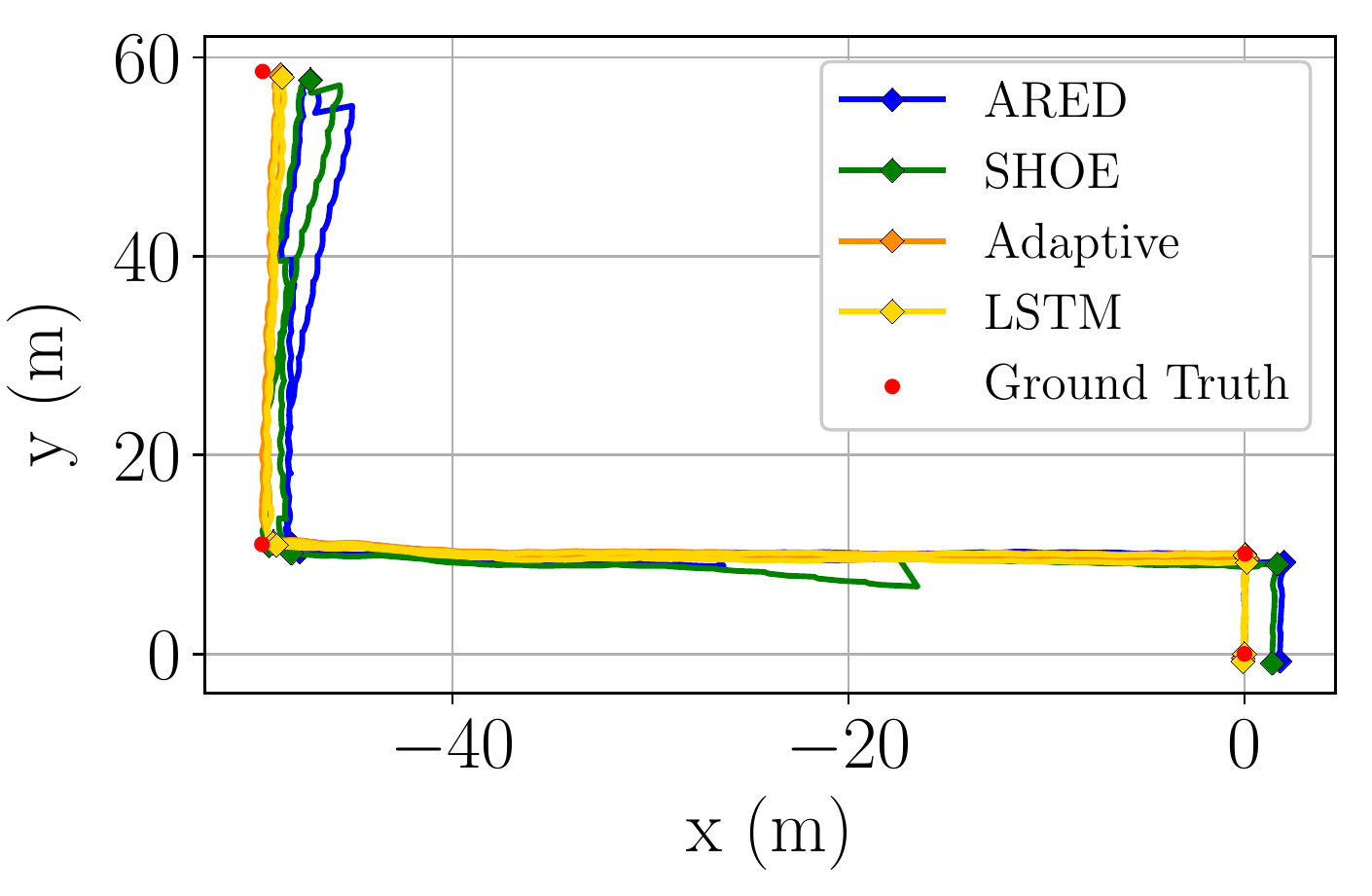}
		\caption{Running trial.}
		\label{fig:hallrun2}
	\end{subfigure} 
	\hfill
	\begin{subfigure}[]{0.32\textwidth}
		\includegraphics[width=\textwidth]{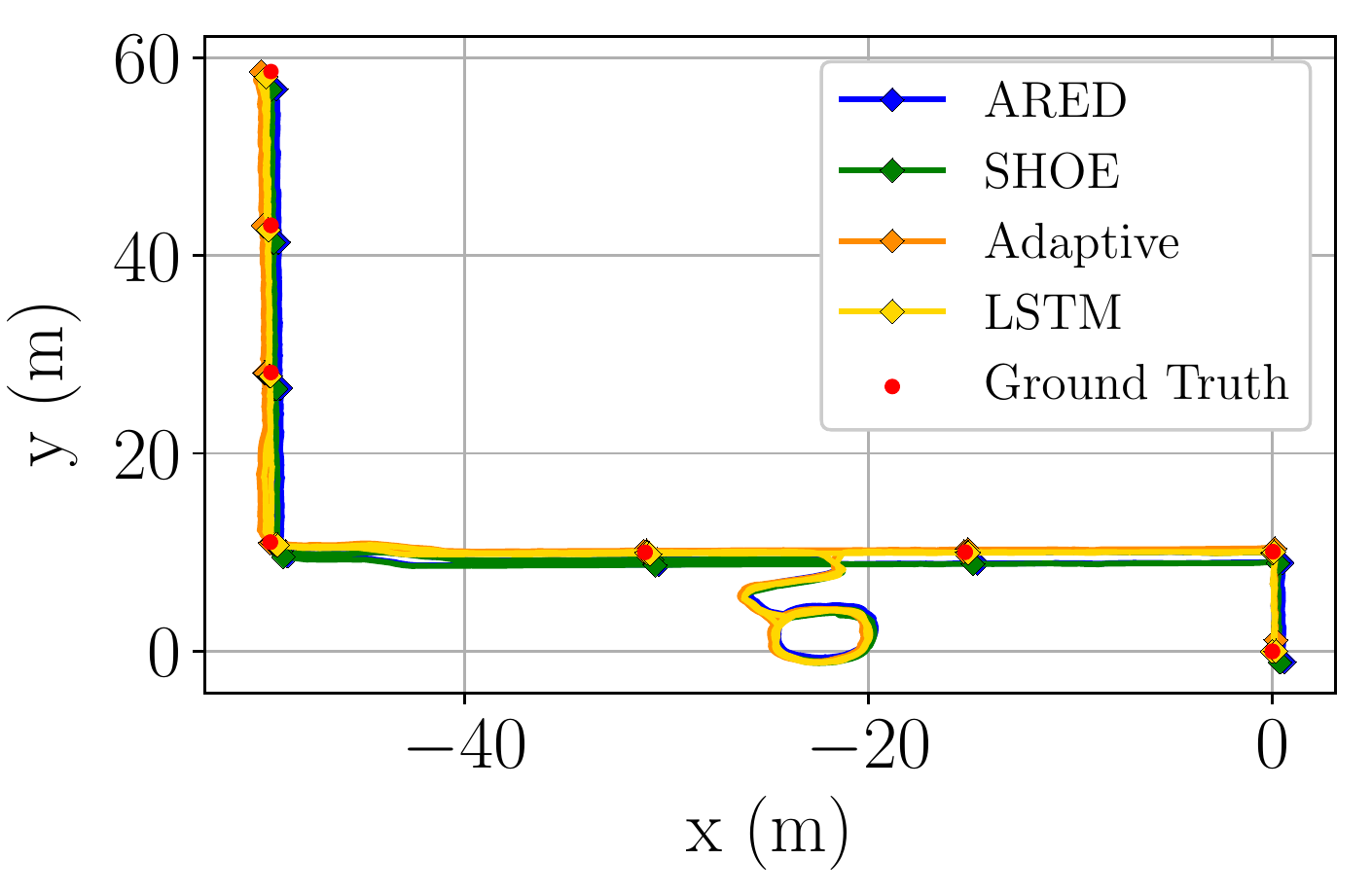}
		\caption{Mixed-motion trial.}
		\label{fig:hallcomb1}
	\end{subfigure}
	\hfill
	\begin{subfigure}[]{0.32\textwidth}
		\includegraphics[width=\textwidth]{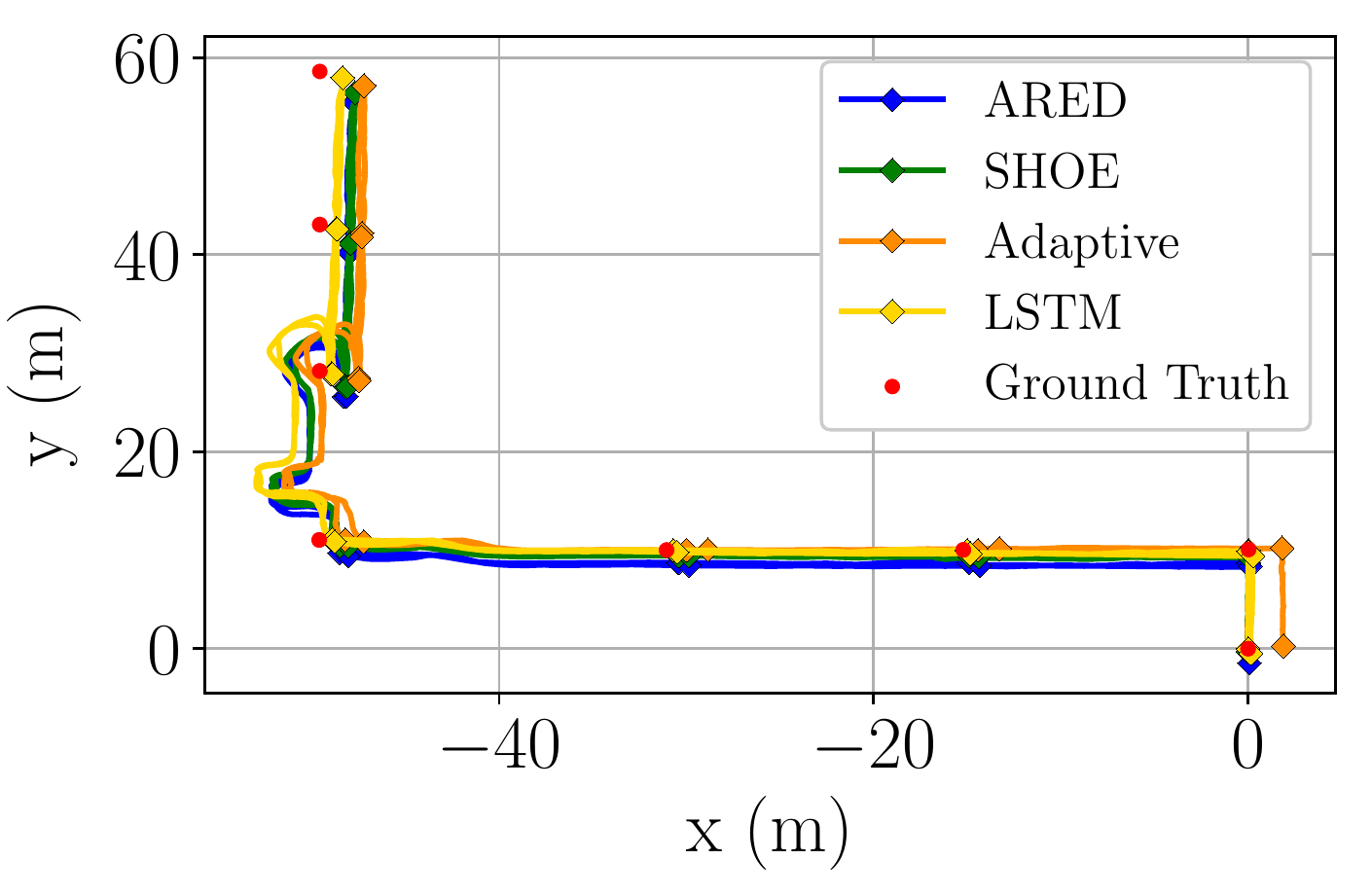}
		\caption{Mixed-motion trial.}
		\label{fig:hallcomb2}
	\end{subfigure}
	\hfill
	\caption{INS trajectories (top-down view) for walking, running, and mixed-motion trials. The proposed data-driven detectors (the adaptive and LSTM-based methods) consistently outperform classical zero-velocity detectors that rely on fixed thresholds.}
	\label{fig:halltraj}
	\vspace{-0.0cm}
\end{figure*}

\subsection{LSTM Training}
\label{sec:lstmtrain}

Our LSTM network is composed of six layers, with 80 units per layer. The final layer is fully connected and reduces the number of outputs from the previous layer (80) to two. A softmax function bounds the two outputs to be within zero and one; the final network output is the argmax of the softmax result, which indicates whether a zero-velocity prediction or an `in motion' prediction is more likely.\footnote{The number of false-positive zero-velocity classifications was reduced by removing all positive events whose confidence fell below 0.85.}

We trained the LSTM-based zero-velocity classifier with the VICON dataset described in  \Cref{tab:datasets}. The motion trials were split into training and validation sets (51 and 9 trials, respectively). We processed the raw, labelled inertial motion data into a form that was appropriate for training: 7000 samples were extracted from each trial, with each sample spanning a window of 100 time steps. An individual sample, $\mbf{x}_i \in\mathds{R}^{100\times6}$, was paired with the zero-velocity label $y_i \in \{0,1\}$ that indicated if the IMU was stationary at the final time step in the window. To ensure that the network was able to generalize to new inertial data, we applied two data augmentation techniques: (1) we rotated each training sample by a `small' random rotation and (2) we scaled each sample by a random factor close to unity. These methods simulated angular velocity and linear acceleration readings from different IMU orientations and motions, reducing overfitting.

A forward pass of $\mbf{x}_i$ through the LSTM propagated the hidden memory state over the 100 time steps. We employed a standard cross-entropy loss function to compare the predicted output, $p_i$, with the target, $y_i$,
\begin{equation}
\mathscr{L}(\mbf{y},\mbf{p}) = -\frac{1}{N}\sum_{i=1}^N y_i\log(p_i) + (1-y_i)\log(1-p_i).
\end{equation}

We implemented our LSTM network in PyTorch \cite{paszke:2017} and trained for 300 epochs using the Adam optimizer \cite{Kingma:2014} on an NVIDIA Titan X GPU. At test time, we achieved real-time performance on a CPU. We refer the reader to \cite{Wagstaff:2018} for more details about the training process.  

\begin{table*}
	\centering
	\footnotesize
	\caption{Position error results (3D ARMSE) from the hallway trials, comparing a fixed-threshold SHOE detector and ARED against our motion-adaptive zero-velocity detector and LSTM-based zero-velocity classifier. The RMSE was computed at several locations in the hallway where floor markers were placed. The location of each marker was surveyed using a Leica Nova MS50 MultiStation.}
	\label{tab:hall-results}
	\begin{threeparttable}	
		\begin{tabular}{cccccccccc}
			\toprule
			\multicolumn{2}{c}{\textbf{Motion}} & \multicolumn{8}{c}{\textbf{ARMSE (m)}} \\ \midrule
			Type & Subject &  & ARED &  &  & SHOE &  & Motion-Adaptive & LSTM \\ \midrule
			&  & \begin{tabular}[c]{@{}c@{}}$\gamma_{walk}$\\ (0.3)\end{tabular} & \begin{tabular}[c]{@{}c@{}}$\gamma_{mid}$\\ (0.55)\end{tabular} & \begin{tabular}[c]{@{}c@{}}$\gamma_{run}$\\ (0.8)\end{tabular} & \begin{tabular}[c]{@{}c@{}}$\gamma_{walk}$\\ ($1\times10^7$)\end{tabular} & \begin{tabular}[c]{@{}c@{}}$\gamma_{mid}$\\ ($8.5\times10^7$)\end{tabular} & \begin{tabular}[c]{@{}c@{}}$\gamma_{run}$\\ ($3.5\times10^8$)\end{tabular} & \begin{tabular}[c]{@{}c@{}}$\gamma_{walk}\,\mathbf{|}\,\gamma_{run}\,\mathbf{|}\,\gamma_{stair}$\\ ($1\times10^7\,\mathbf{|}\,3.5\times10^8\,\mathbf{|}\,1\times10^7$)\end{tabular} & --- \\ \midrule
			 & 0 & 1.11 & 2.93 & 4.25 & 0.54 & 1.26 & 6.16 & 0.56 & \textbf{0.52} \\
			& 1 & 1.10 & 1.35 & 1.57 & 0.84 & 1.15 & 2.09 & 0.85 & \textbf{0.81} \\
\textbf{Walking} ($\sim$220 m) & 2 & 2.51 & 2.73 & 2.95 & 2.27 & 2.66 & 3.86 & 2.43 & \textbf{2.04} \\
			& 3 & \textbf{1.07} & 1.39 & 1.69 & 1.09 & 1.20 & 3.31 & 1.08 & 1.10 \\
			& 4 & 0.65 & 0.73 & 0.82 & \textbf{0.62} & 0.73 & 1.24 & 0.75 & 0.64 \\ \midrule
			& 0 & 1.17 & 1.15 & 2.66 & 3.31 & 0.87 & 2.44 & 0.89 & \textbf{0.87} \\
			& 1 & 3.70 & 1.86 & 1.93 & 16.45 & 1.64 & 0.88 & 0.76 & \textbf{0.67} \\
\textbf{Running} ($\sim$220 m) & 2 & 1.47 & 1.31 & 1.43 & 1.64 & 1.32 & 2.43 & 1.36 & \textbf{0.86} \\
			& 3 & \textbf{0.62} & 1.33 & 1.54 & 1.04 & 1.29 & 1.31 & 0.67 & 0.62 \\
			& 4 & 1.25 & 15.01 & 15.91 & 1.29 & 1.17 & 2.04 & 1.96 & \textbf{0.93} \\ \midrule
			& 0 & 1.15 & 1.72 & 2.42 & 1.05 & 1.25 & 3.59 & 1.00 & \textbf{0.90} \\
			& 1 & 2.29 & 2.02 & 2.20 & 2.00 & 1.73 & 2.31 & 1.35 & \textbf{1.20} \\
\textbf{Combined} ($\sim$240 m) & 2 & 2.47 & 2.97 & 3.30 & 2.09 & 2.75 & 3.91 & 2.35 & \textbf{1.65} \\
			& 3 & 1.11 & 1.28 & 1.37 & 1.10 & 1.22 & 2.37 & 1.10 & \textbf{1.07} \\
			& 4 & 2.57 & 2.66 & 2.64 & 2.51 & 2.63 & 1.60 & \textbf{1.00} & 1.24 \\ \midrule
			\textbf{Mean} &  & 1.52 & 2.48 & 3.03 & 2.38 & 1.44 & 2.81 & 1.14 & \textbf{0.97} \\ \bottomrule
		\end{tabular}
	\end{threeparttable}
	\vspace{-0.2cm}
\end{table*}

\subsection{Hallway Trials}
\label{sec:hallway-results}

We evaluated our proposed data-driven zero-velocity detectors over longer trajectories by carrying out a series of motion trials in the hallways of our building at the University of Toronto Institute for Aerospace Studies (UTIAS). For these experiments, we used a VectorNav VN-100 IMU operating at 200 Hz. Ground truth position information was obtained by surveying the locations of a series of flat markers on the floor using a Leica Nova MS50 MultiStation. Test subjects were given a handheld trigger that they pressed every time their foot was directly on top of a floor marker---this allowed the INS position estimate at the current time step to be compared with the known marker location. For each hallway trial, we report the ARMSE. We refer the reader to our prior publications \cite{Wagstaff:2017,Wagstaff:2018} for a more detailed description of how  ground truth was established.

Data were collected from five different test subjects. All subjects started and ended at the same position on the test course (see \Cref{fig:hallway}). During each trial, a subject travelled approximately 110 m through three different hallways, turned around, and returned to the origin along the same path. We carried out three different types of trials: walking, running, and combined motion. For the walking and combined motion trials, the test subjects recorded their position at every marker; for the running trials, only the corner markers (i.e., markers one, two, five, and eight) were used so that the test subjects did not need to slow down unnecessarily on straight sections of the course. For the combined motion trials, test subjects alternated between walking and running along the course. The path was also extended to allow subjects to enter two rooms (identified as Room 1 and Room 2 in \Cref{fig:hallway}). In Room 1, there was an open space where the test subjects completed three circular laps while alternating between walking and running.  Subjects ascended six steps to enter Room 2 and then descended along a ramp, at which point they re-entered the last hallway. Subjects repeated each trial three times, which resulted in a total of 45 motion trials that covered approximately 6.6 km.

\begin{figure}
	\centering
	\begin{subfigure}[]{0.75\columnwidth}
		\centering
		\includegraphics[width=\columnwidth]{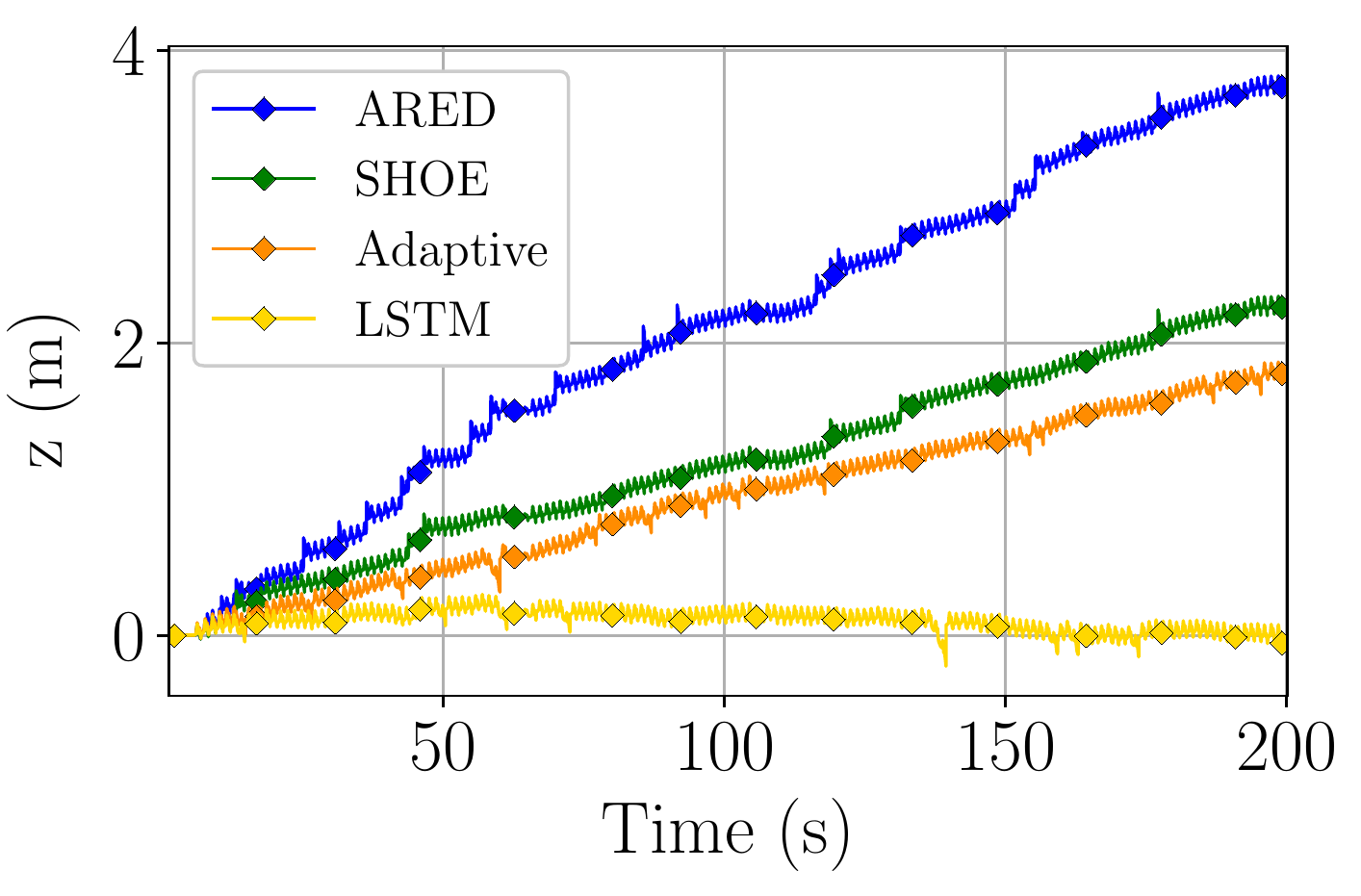}
		\caption{Walking trial (vertical view).}
	\end{subfigure}
	\begin{subfigure}[]{0.75\columnwidth}
		\centering
		\includegraphics[width=\columnwidth]{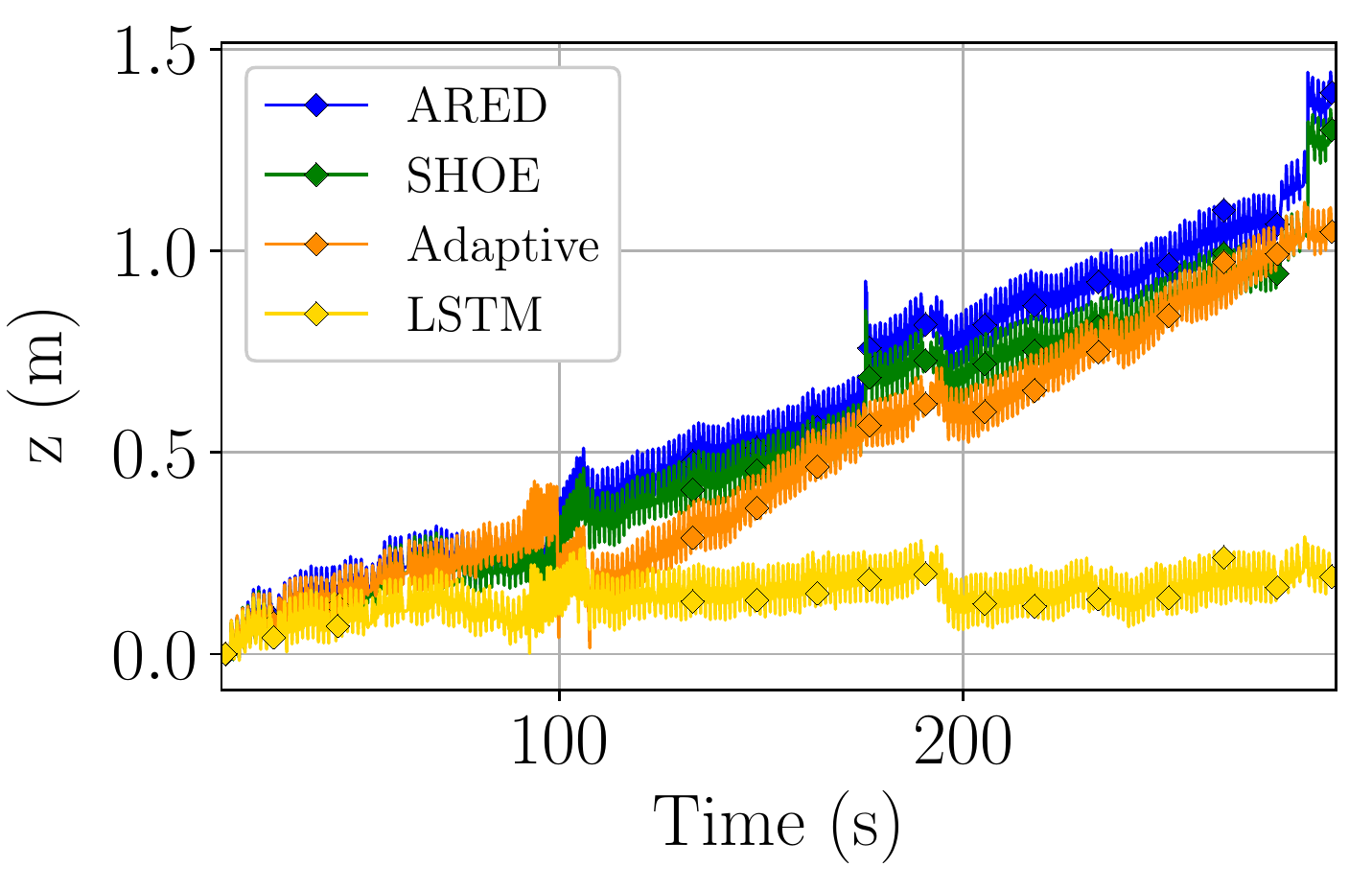}
		\caption{Mixed-motion trial (vertical view).}
	\end{subfigure}
	\hfill
	\caption{Vertical trajectory estimates for motion on a planar surface (a flat floor). The vertical offsets are due entirely to INS drift.}
	\label{fig:hall-vert}
	\vspace{-0.3cm}
\end{figure}

For each trial, we evaluated our baseline INS with the SHOE detector, the ARED, our motion-adaptive detector, and our LSTM-based zero-velocity classifier. \Cref{tab:hall-results} shows the performance of these detectors for the hallway trials. Results for the SHOE detector and the ARED were calculated using three fixed thresholds that were optimized for walking ($\gamma_{walk}$), running ($\gamma_{run}$), and for our VICON test dataset ($\gamma_{mid}$). On average, our LSTM detector achieved a 32.6\% lower ARMSE than the most accurate fixed-threshold detector (SHOE with $\gamma_{mid}$). \Cref{fig:halltraj,fig:hall-vert} illustrate the increase in accuracy that our LSTM-based classifier achieved for the hallway trials. Our motion-adaptive approach produced a similar accuracy, but notably was not as accurate at height estimation (see \Cref{fig:hall-vert}). 

\subsection{Stair-Climbing Trials}
\label{sec:stair-results}

We also collected a unique stair-climbing dataset, acquired from a single test subject who ascended/descended eight flights of stairs within a single stairwell. For each trial, the test subject started and finished at the same 3D position, allowing the 3D loop-closure error to be used as a performance metric. Additionally, the test subject pressed a handheld trigger to indicate when they reached a new flight of steps, which enabled us to compute the vertical error on a per-flight basis. The highest floor was approximately 16.5 m above the ground floor.\footnote{Every flight consisted of 12 steps, each 17.1 cm in height, resulting in a spacing of 2.05 m between floors.}

The test subject performed eight stair-climbing trials in total. The first four trials consisted of ascending from the $1^\text{st}$ to the $3^{\text{rd}}$, $5^{\text{th}}$, $7^{\text{th}}$, and $9^{\text{th}}$ floors, respectively, followed by a return to the ground floor. The last four trials reversed this order, with the wearer starting on the $9^{\text{th}}$ floor and descending to the $7^{\text{th}}$, $5^{\text{th}}$, $3^{\text{rd}}$, and $1^{\text{st}}$ floors, respectively, followed by an ascent back to the $9^{\text{th}}$ floor. In total, 80 flights of stairs were climbed, amounting to 960 steps, and 164.6 m of vertical displacement.

\begin{figure*}[h!]
	\centering
	\begin{subfigure}[t]{0.30\textwidth}
		\includegraphics[width=\textwidth]{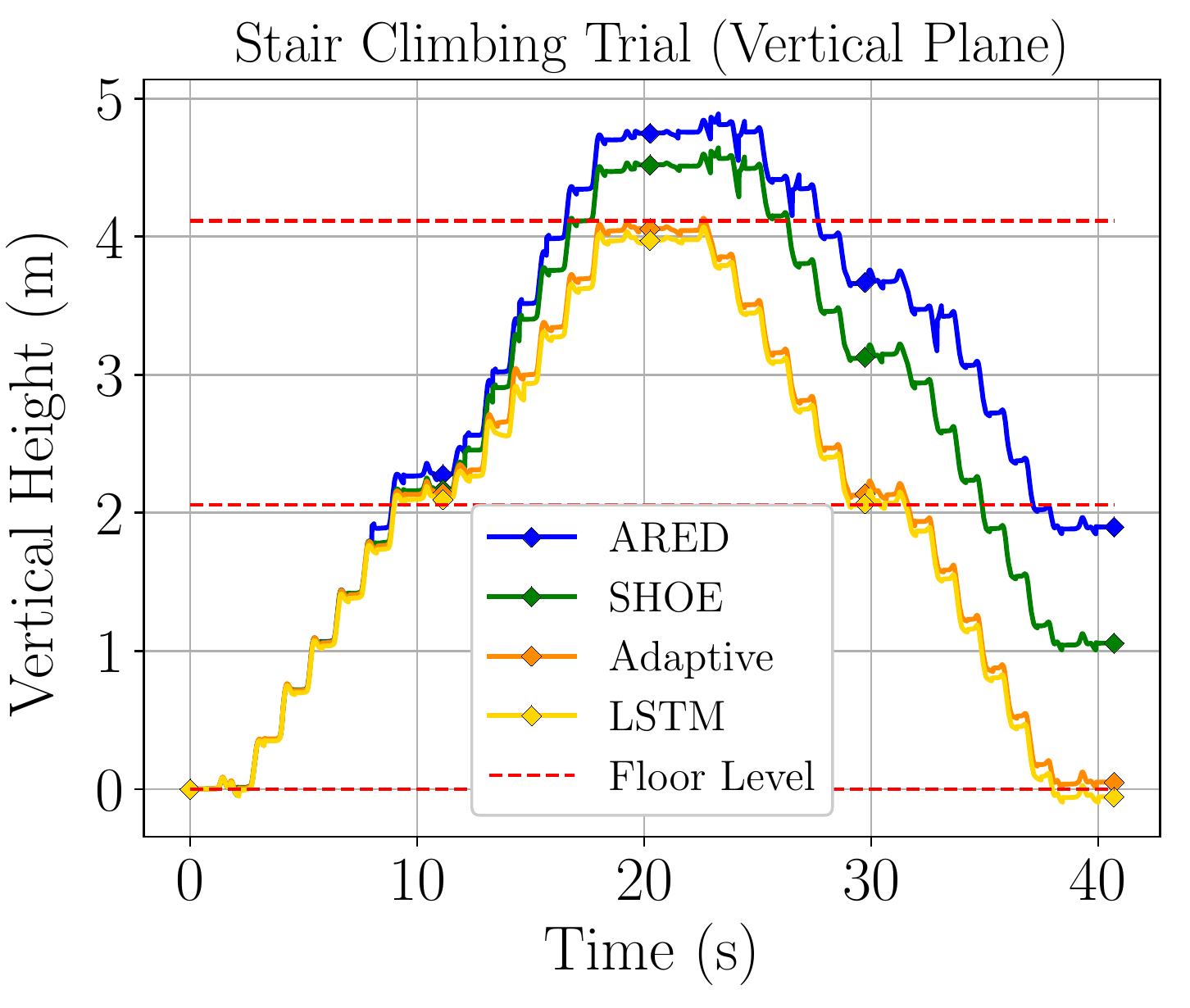}
		\caption{Height over time (2 floors up/down).}
		\label{fig:stairs-vert1}
	\end{subfigure}
	\hfill
	\begin{subfigure}[t]{0.32\textwidth} 
		\includegraphics[width=\textwidth]{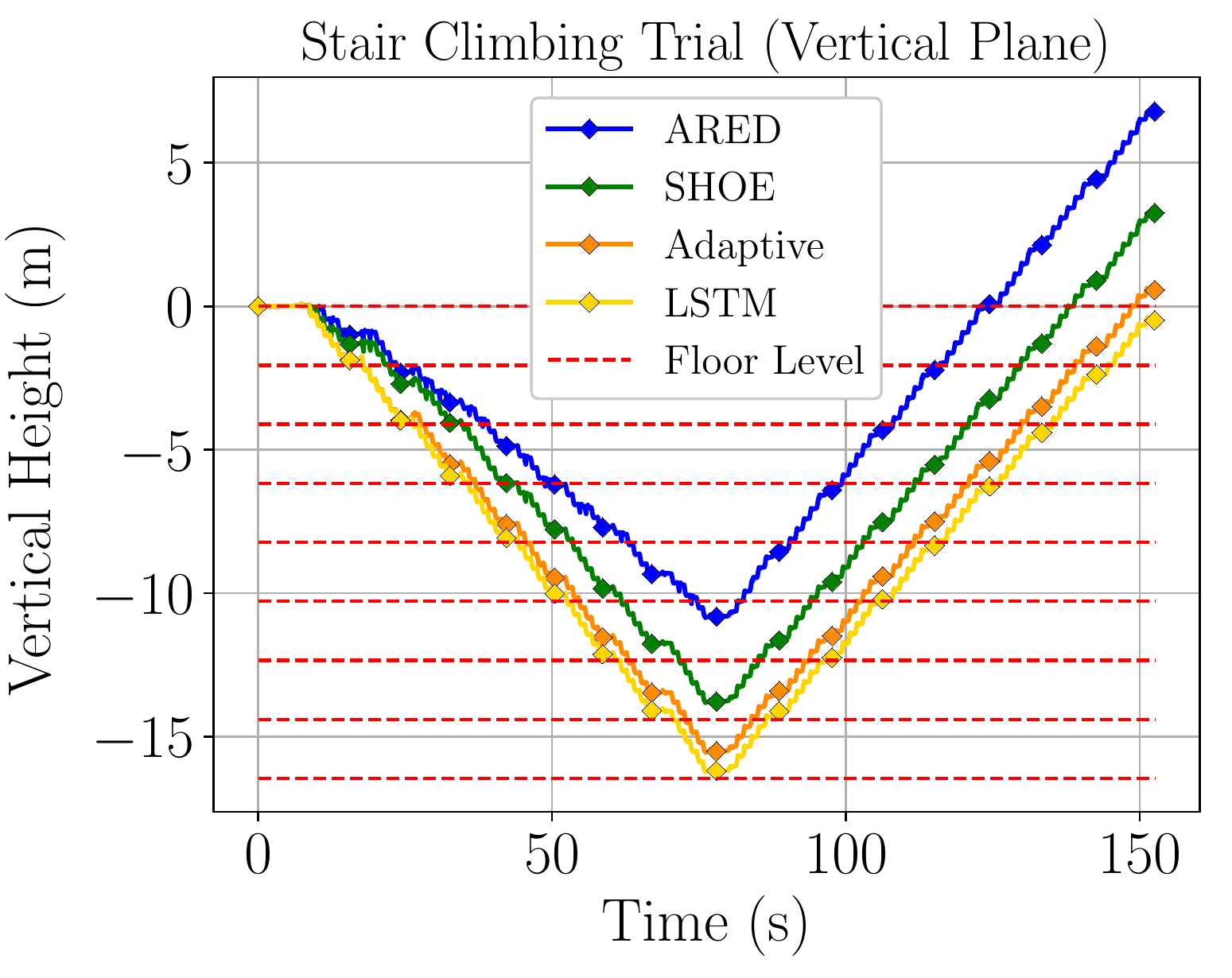}
		\caption{Height over time (8 floors down/up).}
		\label{fig:stairs-vert2}
	\end{subfigure}
	\hfill	
	\begin{subfigure}[t]{0.31\textwidth}
		\includegraphics[width=\textwidth]{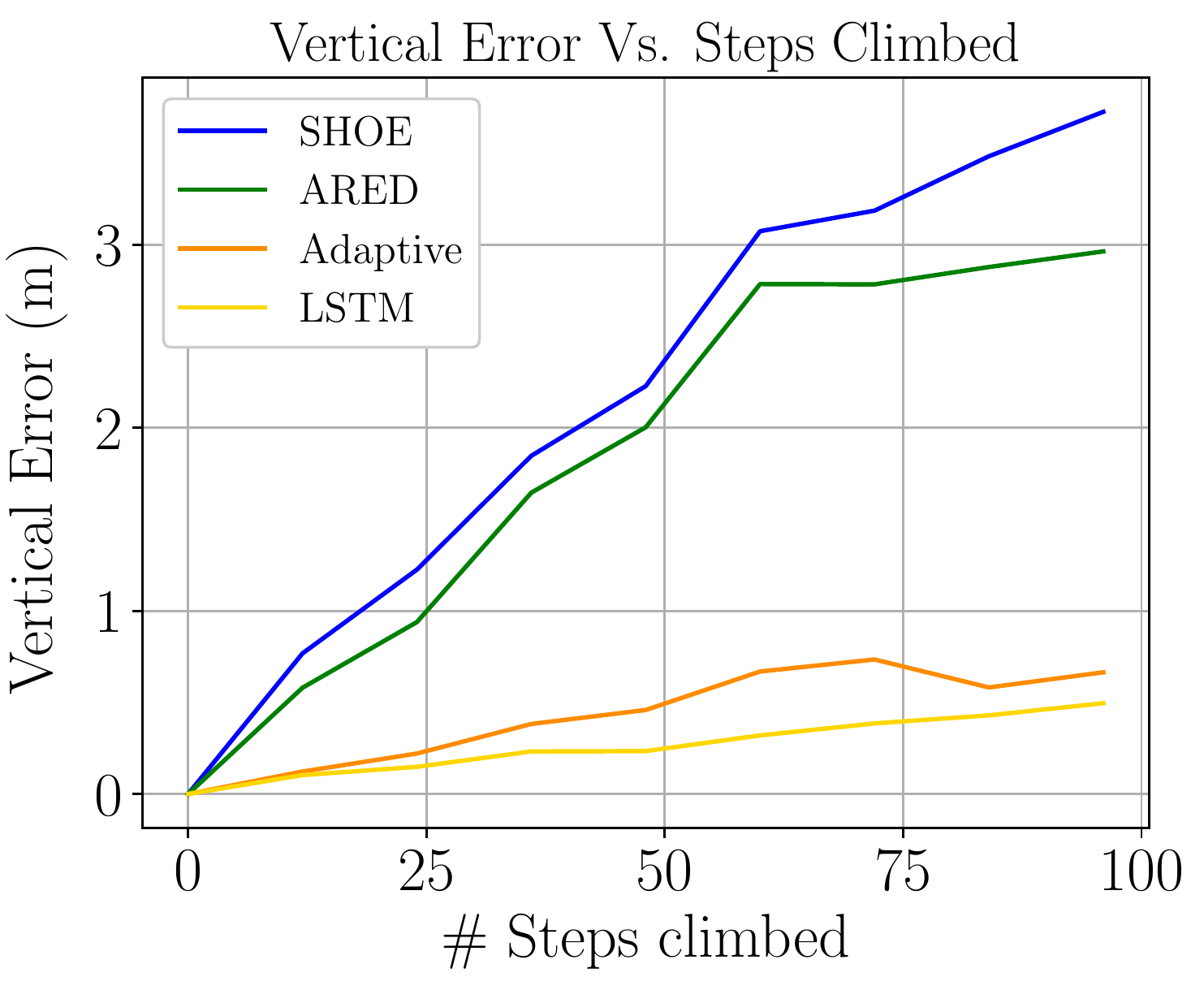}
		\caption{Cumulative error for each detector over all of the stair-climbing trials.}
		\label{fig:error-vs-steps}
	\end{subfigure}
	\caption{Results from the stair-climbing trials. The LSTM-based zero-velocity detector outperformed existing detectors in terms of the vertical position error.}
	\label{fig:stairs}
	\vspace{-0.2cm}
\end{figure*}

Again, we compared the performance of the motion-adaptive detector and the LSTM-based zero-velocity classifier to the SHOE detector and ARED. Both the SHOE detector and the ARED used fixed thresholds in this case, optimized to minimize the error on the VICON test set. While a threshold optimized for stair-climbing would produce more accurate results for the SHOE detector and for the ARED, we omit this from our comparison because optimizing for stair-climbing would substantially degrade detector performance for more common motion types. \Cref{tab:stairs-results} lists the results for the 3D and vertical loop-closure error and the vertical \textit{furthest-point error}, which we define as the difference between the estimated height of the IMU and its known height at the furthest point of the trajectory. In all cases, our data-driven detectors outperformed the SHOE detector and the ARED by a large margin. \Cref{fig:stairs} depicts the results from our stair-climbing experiments. \Cref{fig:stairs-vert1,,fig:stairs-vert2} illustrate failure modes of the SHOE detector and the ARED: these hand-engineered detectors both overestimated \textit{and} underestimated vertical displacement to such a degree that even floor-level accuracy could not be attained. In contrast, our proposed detectors on average produced a vertical estimate that exceeded the accuracy needed for floor-level estimation. \Cref{fig:error-vs-steps} plots the vertical displacement error against the number of steps climbed for all of the trials. The proposed detectors maintained an accuracy of better than one meter over a range of 100 steps, while the SHOE detector and the ARED exceeded this error bound within 25 steps.

\begin{table}[!t]
\centering
\caption{Results from the stair-climbing trials. Loop-closure error is computed at the end of the trajectory (when the wearer returned to the origin).}
\label{tab:stairs-results}
	\begin{threeparttable}
	\begin{tabular}{c@{\hspace{0.9\tabcolsep}}c@{\hspace{0.9\tabcolsep}}c@{\hspace{0.9\tabcolsep}}c@{\hspace{0.9\tabcolsep}}c@{\hspace{0.9\tabcolsep}}}
		\toprule
		\textbf{\# Flights} & \textbf{Detector} & \multicolumn{3}{c}{\textbf{Position Errors (m)}} \\ \midrule
		&  & \begin{tabular}[c]{@{}c@{}} Loop-Closure\\ (3D Error)\end{tabular} & \begin{tabular}[c]{@{}c@{}}Loop-Closure\\ (Vertical Error)\end{tabular} & \begin{tabular}[c]{@{}c@{}}Furthest-Point\\ (Vertical Error)\end{tabular} \\ \midrule
		& ARED & 1.895 & 1.878 & 1.265 \\
		\textbf{2} & SHOE & 1.014 & 1.001 & 0.703 \\
		& Adaptive & \textbf{0.091} & 0.055 & \textbf{0.109} \\
		& LSTM & 0.098 & \textbf{0.052} & 0.115 \\
		&  &  &  &  \\
		& ARED & 3.155 & 2.865 & 2.015 \\
		\textbf{4} & SHOE & 1.841 & 1.18 & 1.526 \\
		& Adaptive & \textbf{0.321} & \textbf{0.199} & \textbf{0.146} \\
		& LSTM & 0.343 & 0.251 & 0.172 \\
		&  &  &  &  \\
		& ARED & 2.443 & 2.109 & 3.294 \\
		\textbf{6} & SHOE & 1.833 & 1.499 & 2.945 \\
		& Adaptive & 1.124 & 0.513 & 0.931 \\
		& LSTM & \textbf{0.362} & \textbf{0.279} & \textbf{0.396} \\
		&  &  &  &  \\
		& ARED & 4.83 & 4.811 & 3.725 \\
		\textbf{8} & SHOE & 2.312 & 2.244 & 2.963 \\
		& Adaptive & 0.853 & 0.648 & 0.665 \\
		& LSTM & \textbf{0.73} & \textbf{0.495} & \textbf{0.496} \\ \midrule
		& ARED & 3.081 & 2.916 & 2.575 \\
		\textbf{Mean} & SHOE & 1.75 & 1.481 & 2.034 \\
		& Adaptive & 0.597 & 0.354 & 0.463 \\
		& LSTM & \textbf{0.384} & \textbf{0.269} & \textbf{0.295} \\ \bottomrule
	\end{tabular}
\end{threeparttable}
\end{table}

\subsection{Discussion}

Our proposed zero-velocity detectors were able to operate consistently with five different test subjects, while being trained with data from a single subject only. Furthermore, the detectors operated largely independently of the orientation of the IMU on the foot, and also were invariant to the location of the IMU on the shoe and the shoe type in general (we did not specify where the IMU should be mounted, or what type of shoe should be worn for data collection).  

The learning-based detectors were able to run on a CPU in real time; we report the operating frequencies in \Cref{tab:compute-time}: both of the proposed detectors, although slower than the classical detectors, were able to function at a frequency above approximately 2 kHz. In general, this operating frequency is much greater than the update frequency of an IMU. 

\begin{table}[!b]
    \vspace{-2mm}
	\centering
	\caption{Operating frequencies for several zero-velocity detectors. Our proposed detectors are capable of running in real time on an Intel i7-6700HQ CPU.}
	\label{tab:compute-time}
	\begin{threeparttable}
		\begin{tabular*}{\columnwidth}{l@{\extracolsep{\fill}}*{5}{c}}
			\toprule
			& \textbf{ARED} & \textbf{SHOE} & \textbf{Adaptive} & \textbf{LSTM} \\ \midrule
			\textbf{\begin{tabular}[c]{@{}c@{}}Operating Frequency\\ (kHz)\end{tabular}} & 215 & 32.2 & 1.97 & 3.35 \\ \bottomrule
		\end{tabular*}
	\end{threeparttable}
\end{table}
 
\begin{figure}[t]
	\small
	\centering
	\includegraphics[width=0.75\columnwidth]{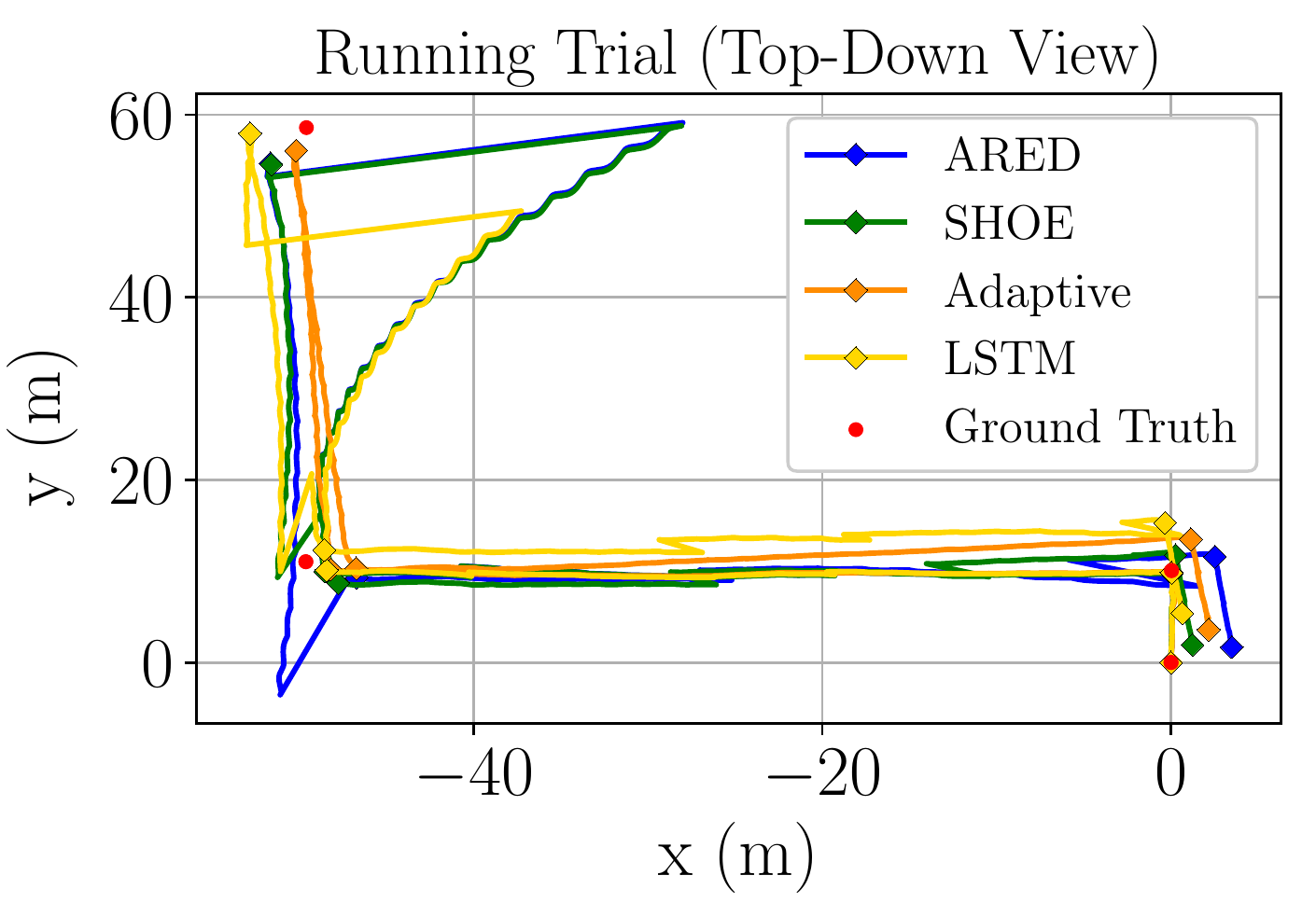}
	\caption{A sample trajectory where our LSTM fails when given out-of-distribution data (the running speed in the hallway was faster than what could be attained within our VICON room).}
	\label{fig:lstm-breakdown}
\end{figure}

Despite the increase in accuracy provided by the learning-based detectors, there are areas where further improvements could be made. First, the motion classifier could be trained with inertial data that is representative of new motion types (such as crawling) where the optimal zero-velocity threshold is very different than for other motions. Second, the LSTM-based zero-velocity classifier could make use of training data collected in areas other than our VICON room and at greater velocities (as seen in \Cref{fig:lstm-breakdown}, the LSTM model was occasionally unable to detect zero-velocity events when a wearer's speed was outside of the training distribution). 

Lastly, we note that the increase in positioning accuracy derived from the use of our learning-based detectors is primarily a result of an improved velocity estimate. Although the zero-velocity updates do impact the IMU roll and pitch estimates, the yaw (heading) remains unobservable \cite{Nilsson:2014}. Classical zero-velocity detectors such as the SHOE detector permit roll and pitch to be recovered accurately (because the magnitude of the gravity vector, which appears in the error-state computation, is large compared to other measured accelerations); we have found that the use of learning-based detectors does not substantially change the accuracy of the attitude estimate.

\section{Generalizing to New IMUs}
\label{sec:domain-adapt}

Although we have shown that our learned LSTM-based zero-velocity classifier generalizes to new IMU placements, new wearers, and varying motion types, we do not expect it to function properly with an IMU that has significantly different \emph{hardware} characteristics. To improve this, we propose a data manipulation technique that transforms the data within the training dataset to be representative of the outputs of a different IMU. By retraining the zero-velocity classifier with the transformed data, the LSTM network is better able to generalize to inertial data from a different, lower-quality sensor.

\subsection{IMU Data Transformation}

Our proposed transformation technique adds zero-mean Gaussian noise to each IMU channel and then downsamples the original data to a lower frequency. These steps are meant to account for varying IMU sample rates and for varying measurement quality. The newly-transformed data are then used to retrain the LSTM network to be compatible with an alternate IMU. Importantly, this approach obviates the need to collect any new training data with the alternate IMU, which can be tedious, time-consuming, and may not always be possible.

We note that our data manipulation approach can only be used to simulate an IMU with higher levels of noise and a lower update frequency that the IMU used to collect the training data. Despite this limitation, given a training dataset collected with a high-quality IMU (e.g., our VICON dataset), there is a wide array of lower-cost IMUs to which the data manipulation technique is applicable. 

\subsection{IMU Generalization Experiments}

Our LSTM-based zero-velocity detector was initially trained with data from the VICON dataset, collected with the MicroStrain 3DM-GX3-25 IMU operating at 200 Hz. While we demonstrated in \cite{Wagstaff:2018} that the learned model generalized to the (comparable) VectorNav VN-100 IMU at test time, both IMUs have similar noise characteristics and operate at the same frequency. Herein, as a proof-of-concept study, we show that our model can be retrained to operate with the low-cost Osmium MIMU22BTP IMU \cite{Gupta:2015}, running at 125 Hz, which has significantly noisier accelerometers and gyroscopes. For brevity, we refer to this low-cost IMU as the Osmium IMU.

We transformed the VICON training dataset to be representative of the data generated by the Osmium by downsampling from 200 Hz to 125 Hz\footnote{We applied a first-order low-pass (Butterworth) filter with a cutoff frequency of 40 Hz to prevent any aliasing effects.} and adding Gaussian noise to each sample ($\sigma_{a}=1\times10^{-2}$ and $\sigma_{\omega}=1.74\times10^{-3}$ for the accelerometer and gyroscope channels, respectively). We then retrained the LSTM model with the modified VICON dataset using the same training procedure described in \Cref{sec:lstm-zv}. 

We evaluated the LSTM classifier (i.e., both the original and the retrained models) on the Hallway 2 dataset (see \Cref{tab:datasets}) that included data from five test subjects walking and running with the Osmium mounted on their right foot. To be consistent with prior work that used the Hallway 2 dataset \cite{Wagstaff:2017}, we evaluated the end-point error at the furthest point along the trajectory and we report the average end-point error for each motion type. \Cref{tab:domain-adapt} shows that the retrained LSTM-based zero-velocity classifier resulted in the most accurate position estimates---its use led to an 18.4\% reduction in mean error with respect to the original LSTM network trained with the MicroStrain data. These results indicate that the proposed IMU generalization method can facilitate the use of the LSTM-based classifier with new IMUs without the need to recollect any training data.

\begin{table}[]
	\centering
	\caption{IMU generalization results. When evaluating the LSTM on the Hallway 2 dataset (which was collected with the low-cost Osmium IMU), the retrained LSTM network outperformed the original LSTM that was trained with the MicroStrain IMU data.}
	\label{tab:domain-adapt}
	\begin{tabular}{c @{\hspace{0.1\tabcolsep}} c @{\hspace{0.1\tabcolsep}} c @{\hspace{0.1\tabcolsep}} c @{\hspace{0.1\tabcolsep}} c @{\hspace{0.1\tabcolsep}} c @{\hspace{0.5\tabcolsep}} c @{\hspace{0.5\tabcolsep}} c @{\hspace{0.5\tabcolsep}} }
		\toprule
		\multicolumn{2}{c}{\textbf{Motion}} & \multicolumn{6}{c}{\textbf{End-Point ARMSE (m)}} \\ \midrule \T \T \B
		Type & Subject & \multicolumn{3}{c}{SHOE} & Adaptive & \begin{tabular}[c]{c@{}c@{}} Original \\ LSTM \end{tabular} & \begin{tabular}[c]{@{}c@{}}Retrained\\ LSTM\end{tabular} \\ \midrule \T \T \B
		&  & \multicolumn{1}{c}{$\gamma_{walk}$} & \multicolumn{1}{c}{$\gamma_{mid}$} & $\gamma_{run}$ &  &  &  \\ \midrule \T \T \B
		\multirow{5}{*}{Walking} & 1 & \multicolumn{1}{c}{1.17} & \multicolumn{1}{c}{1.86} & 2.23 & 1.24 & \textbf{0.98} & 1.30 \\
		& 2 & \multicolumn{1}{c}{0.76} & \multicolumn{1}{c}{1.10} & 1.27 & 0.74 & 0.77 & \textbf{0.72} \\
		& 3 & \multicolumn{1}{c}{\textbf{0.31}} & \multicolumn{1}{c}{0.65} & 0.85 & 0.34 & 0.45 & 0.32 \\
		& 4 & \multicolumn{1}{c}{0.60} & \multicolumn{1}{c}{1.47} & 3.07 & 0.62 & \textbf{0.48} & 0.56 \\
		& 5 & \multicolumn{1}{c}{1.02} & \multicolumn{1}{c}{1.73} & 2.17 & 1.01 & 1.26 & \textbf{0.92} \\ \midrule \T \T \B
		\multirow{5}{*}{Running} & 1 & \multicolumn{1}{c}{12.36} & \multicolumn{1}{c}{2.39} & 2.39 & \textbf{2.22} & 2.30 & 2.40 \\
		& 2 & 99.11 & 3.74 & 1.82 & 1.82 & 4.02 & \textbf{1.72} \\
		& 3 & 62.27 & \textbf{1.73} & 1.82 & 1.82 & 1.79 & 1.80 \\
		& 4 & 154.88 & 3.37 & 4.90 & 4.89 & 3.31 & \textbf{3.28} \\
		& 5 & 128.31 & \textbf{1.12} & 1.48 & 1.48 & 4.09 & 1.33 \\ \midrule \T \T \B
		\multirow{5}{*}{Combined} & 1 & 5.45 & 2.61 & 2.91 & 2.70 & 2.69 & \textbf{2.60} \\
		& 2 & 2.80 & 2.33 & 2.40 & 2.30 & \textbf{1.93} & 2.00 \\
		& 3 & 2.50 & 0.76 & 0.86 & \textbf{0.70} & 0.71 & 0.72 \\
		& 4 & 11.66 & 2.26 & 2.72 & 2.48 & 2.13 & \textbf{2.12} \\
		& 5 & 0.87 & 1.29 & 1.64 & 1.09 & \textbf{0.77} & 0.83 \\ \midrule \T \T \B
		Mean &  & 32.27 & 1.89 & 2.17 & 1.70 & 1.85 & \textbf{1.51} \\ \bottomrule
	\end{tabular}
	\vspace{-0.2cm}
\end{table}

\section{Conclusion}

Ensuring correct zero-velocity detection is a crucial step towards achieving accurate indoor localization from foot-mounted inertial data. We have presented two new techniques for zero-velocity detection that are robust to varying motion type. Our SVM-based motion-adaptive detector operates as a motion classifier to actively update the zero-velocity threshold of a classical zero-velocity detector. In contrast, our LSTM-based zero-velocity classifier directly outputs zero-velocity pseudo-measurements without requiring any motion-specific parameter tuning. We showed that both of our proposed techniques outperform existing threshold-based detectors on several large datasets from different individuals performing locomotion tasks and that the LSTM-based zero-velocity classifier produced the lowest average error on datasets incorporating stair-climbing motions. We also introduced a generalization method that permits the use of our LSTM-based classifier with lower-cost IMUs without the need to collect additional training data. Lastly, we have made our datasets and our Python-based INS publicly available---we invite others to benchmark their implementations against our own.

As future work, we intend to investigate the end-to-end training of probabilistic measurement functions for zero-velocity-aided inertial navigation in a fully learnable Bayesian filtering framework, similar to \cite{Haarnoja:2016}.

\section*{Acknowledgment}
This work was supported in part by the Natural Sciences and Engineering Research Council (NSERC) of Canada. We gratefully acknowledge the contribution of NVIDIA Corporation, who provided the Titan X GPU used for this research through their Hardware Grant Program.

\bibliographystyle{IEEEtran}
\bibliography{2019-wagstaff-robust.bib}

\end{document}